\documentclass[10pt,twocolumn,letterpaper]{article}

\usepackage[pagenumbers]{cvpr}

\usepackage{adjustbox}
\usepackage{makecell}
\usepackage{wrapfig}
\usepackage{etoolbox}
\newlength{\wrapfigsep}
\makeatletter
\patchcmd{\WF@putfigmaybe}{\columnsep}{\wrapfigsep}{}{}
\patchcmd{\WF@putfigmaybe}{\columnsep}{\wrapfigsep}{}{}
\patchcmd{\WF@putfigmaybe}{\columnsep}{\wrapfigsep}{}{}
\makeatother
\usepackage{algorithm}
\usepackage{algpseudocode}

\newcommand{\method}{GEAR\xspace}
\newcommand{\projectpage}{\url{https://zju3dv.github.io/geometry-as-address/}}

\definecolor{myPurple}{rgb}{0.4, .0, .8}
\definecolor{myGreen}{rgb}{0, 0.6, .3}
\definecolor{myRed}{rgb}{0.8, .2, .2}
\definecolor{myOrange}{rgb}{0.8, 0.45, 0.0}
\definecolor{myBlue}{rgb}{.0, .0, 1.0}
\definecolor{myBlue2}{rgb}{.0, 1.0, 1.0}
\definecolor{myBlack}{rgb}{.0, .0, 0.0}
\definecolor{darkmidnightblue}{rgb}{0.0, 0.2, 0.4}
\definecolor{MyGreen}{rgb}{0.02,0.5,0.02}

\definecolor{cvprblue}{rgb}{0.21,0.49,0.74}
\usepackage[pagebackref,breaklinks,colorlinks,allcolors=cvprblue]{hyperref}

\def\paperID{*****}
\def\confName{CVPR}
\def\confYear{2026}

\renewcommand{\abstract}{%
   \iftoggle{cvprpagenumbers}{}{\thispagestyle{empty}}%
   \centerline{\large\bf Abstract}%
   \vspace*{4pt}\noindent%
   \it\ignorespaces%
}

\title{Geometry as Address: Routing Attention to Visual Memory for Long-Horizon Camera-Controlled Video Generation}

\author{
Zesong Yang$^{1}$ \quad
Weikai Chen$^{3\ddagger}$ \quad
Liyuan Cui$^{1}$ \quad
Lutao Jiang$^{2}$ \quad
Runze Zhang$^{3}$ \\
Yingda Yin$^{3}$ \quad
Xiaoyang Huang$^{3}$ \quad
Kai Yan$^{3}$ \quad
Keyang Luo$^{3}$ \\
Wangguandong Zheng \quad
Xin Wang$^{3}$ \quad
Hujun Bao$^{1}$ \quad
Zhaopeng Cui$^{1\dagger}$ \\[0.5em]
$^{1}$State Key Laboratory of CAD\&CG, Zhejiang University \quad
$^{2}$HKUST(GZ) \quad
$^{3}$LIGHTSPEED
}

\begin{document}

\twocolumn[{%
\renewcommand\twocolumn[1][]{#1}%
\maketitle

\setlength{\abovedisplayskip}{0.275 em}
\setlength{\belowdisplayskip}{0.275 em}

\begin{center}
    \centering
    \captionsetup{type=figure}
    \vspace{-1.0 em}
    \includegraphics[width=0.98\textwidth]{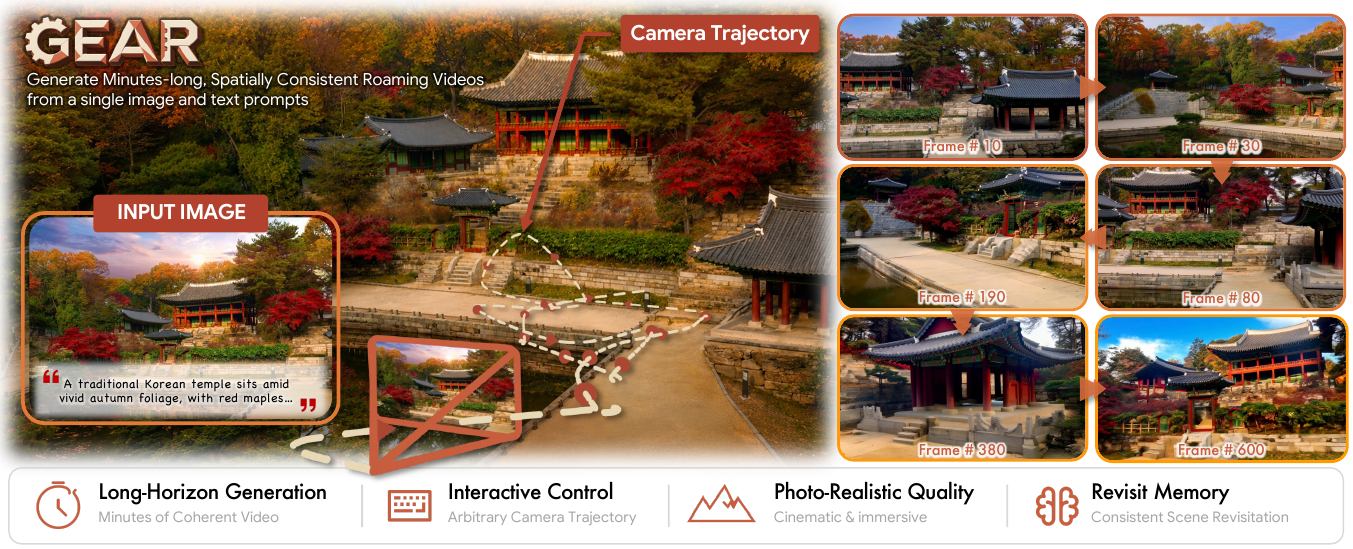}
    \caption{
    \textbf{Precise camera control and persistent scene memory for long-horizon exploration.}
    Given a single image and a text prompt, \textbf{\method} generates minute-long, photorealistic videos along arbitrary user-specified camera trajectories, while preserving spatial consistency and faithfully recovering previously observed content upon revisitation.
    }
    \label{fig:teaser}
    \vspace{1.5 em}
\end{center}%
}]

{
\renewcommand{\thefootnote}{}
\footnotetext{$^{\ddagger}$Project Lead. \quad $^{\dagger}$Corresponding Author.}
}

\begin{abstract}
    Long-horizon camera-controlled video generation requires recovering previously observed content from an ever-growing visual history. Existing approaches either search historical context implicitly or reconstruct it into persistent 3D memory, facing inefficient memory access or accumulated geometric errors.
    Our key insight is that geometry need not explain the scene -- it only needs to determine \emph{where visual memory should be read from}, while attention decides \emph{what should be recovered}.
    Based on this insight, we introduce \method, a Geometry-Enabled Attention Routing framework that uses geometry as an explicit token-level address for visual memory.
    Rather than fusing historical observations into a persistent global 3D representation, \method retains them as frame latents and uses per-frame geometry only to establish token-level correspondences with target views, thereby avoiding persistent error accumulation from global fusion.
    Guided by these correspondences, a proposed Geometric Correspondence Attention (GCA) selectively injects geometrically matched historical features into target noisy patches during denoising.
    We further introduce an Invisible Octree to accumulate visibility evidence and reject geometrically plausible but occluded correspondences.
    Extensive experiments demonstrate that \method achieves state-of-the-art visual quality, precise camera control, and revisit consistency, enabling minute-long video generation along challenging trajectories.
    Additional videos are available at \projectpage.
    \end{abstract}

\begin{figure*}[t]
    \centering
    \scriptsize
    \includegraphics[width=0.975\textwidth]{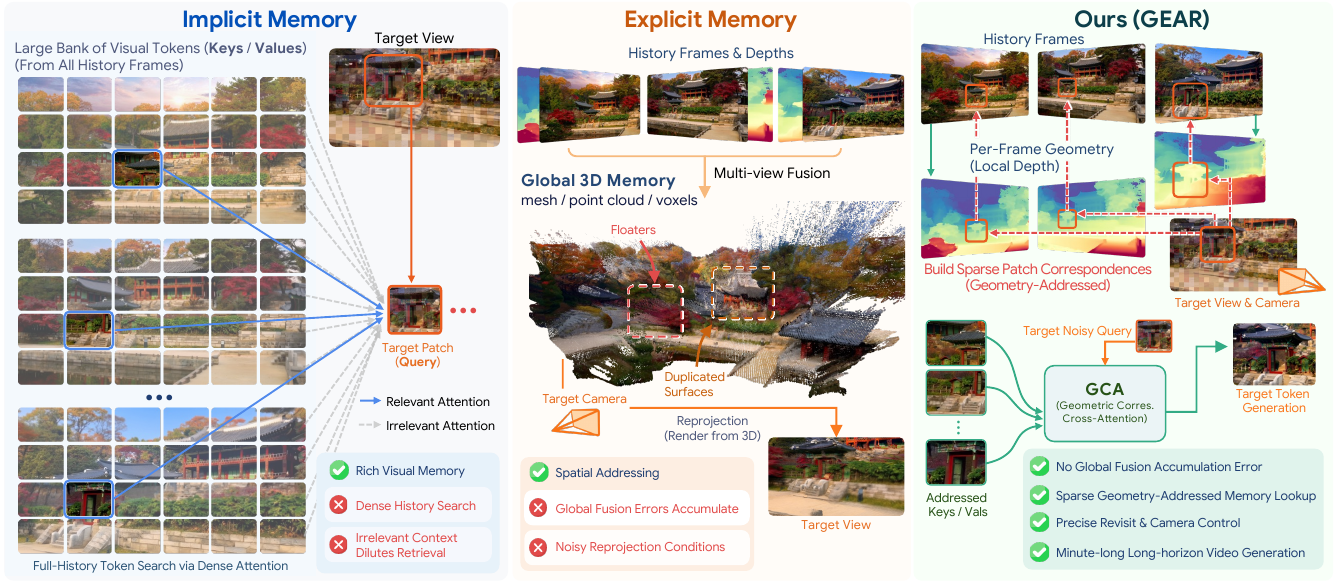}
    \caption{
        \textbf{Comparison of long-term memory paradigms.}
        Implicit memory preserves rich visual history but requires dense search over an increasingly large token set, while explicit 3D memory provides spatial addressing at the cost of accumulated reconstruction and fusion errors. \method instead uses per-frame geometry only to address historical latent patches, enabling sparse, spatially grounded memory retrieval without persistent global 3D fusion.
    }
    \label{fig:motivation}
\end{figure*}

\section{Introduction}

Recent advances in video generation have enabled realistic and controllable visual synthesis~\citep{wan2025wan, seedance2026seedance2, klingai2026kling3, googledeepmind2025veo3}, opening the door to interactive world generation where camera trajectories or user actions control the synthesized observations~\citep{team2026alayaworldv11,zhu2026astra,team2026advancing,hyworld2026,alibaba2026happyoyster,parkerholder2025genie3,mao2025yume15textcontrolledinteractiveworld}.
Extending these models to long-horizon exploration, however, requires more than generating plausible short clips: the model must accurately follow the prescribed camera trajectory while preserving previously observed scene content over temporal gaps.
When a scene region is revisited after leaving the model's temporal context, its appearance must be recovered from past observations rather than inferred from the current context.
Long-horizon generation therefore becomes a \textit{memory access problem}: for each target region, the model must identify and retrieve the relevant visual evidence from history.

Existing approaches address this problem through two paradigms, illustrated in Fig.~\ref{fig:motivation}.  History-based implicit methods retain previously generated observations as context memory and recover relevant information through attention~\citep{he2024cameractrl,bai2025recammaster,xiao2026worldmem,li2026cameras,yu2024gamefactory,xiao2026worldmem,yu2025cam,sun2025worldplay}.
This preserves rich visual information, but as history grows, the model must search over an increasingly large token set to recover the few observations relevant to the current viewpoint, making memory access costly and vulnerable to irrelevant context.
Frame retrieval or history compression~\citep{xiao2026worldmem,yu2025cam,zhang2025framepack,wu2026infiniteworld} alleviates this burden, but operates at a coarser granularity or may discard information needed for precise revisitation.
Reconstruction-based explicit methods instead fuse historical observations into a global 3D representation and reproject it to target views~\citep{ren2025gen3c,wu2025video,zhao2026spatia,chen2026anyrecon}. While this provides natural spatial addressing, local geometry errors can become persistent after global fusion and propagate into subsequent generations through noisy reprojection.

Recent methods have begun to bridge these two paradigms by exploiting geometry without relying on a single globally fused 3D memory.
AnchorWeave~\citep{wang2026anchorweave} maintains multiple local geometric representations,
reprojects retrieved memories as target-view anchor videos and adaptively fuses them through ControlNet~\citep{zhang2023adding}.
UCM~\citep{xu2026ucm} instead warps positional encodings to geometrically align interactions between historical and target tokens.
Most closely related, Lyra 2.0~\citep{shen2026lyra} warps source coordinates and depth into target views, then injects their embeddings into DiT tokens.
Despite these advances, geometry is still used to \emph{select, align, or construct geometric conditioning signals from history}, while \emph{access to the underlying visual features remains indirect}. This motivates a more explicit separation between geometry and visual memory: rather than using correspondence only to guide generation, we use it to directly determine which historical visual tokens each target token can access.

Rather than asking geometry to explain the scene, our key insight is to let geometry answer only \textbf{where memory should be read from}, while leaving \textbf{what should be recovered} to visual attention over the matched historical features.
Based on this insight, we introduce \textbf{\method}, a Geometry-Enabled Attention Routing framework that \textbf{uses geometry as an explicit token-level address for visual memory}.
For each target token, \method identifies a sparse set of geometrically matched historical tokens and gathers their visual features as keys and values for attention. In this way, geometry constrains the memory search space, while visual attention resolves which historical evidence is most useful for generation.
We realize this mechanism through Geometric Correspondence Attention (GCA), which exposes each noisy target token only to its geometrically matched historical memory tokens as keys and values, and injects the aggregated features through a residual branch during denoising.
Geometry thus serves as a transient address rather than persistent scene state: it explicitly determines where each target token can retrieve visual evidence, while correspondence errors remain local to individual memory accesses instead of accumulating across views.

Cross-view projection alone, however, may produce false correspondences when a source-visible surface becomes occluded in the target view.
We therefore introduce an Invisible Octree that accumulates visibility evidence over time and filters such correspondences without storing scene appearance.
Together, these designs enable efficient patch-level memory access over long trajectories:
rather than requiring the video model to search the entire history or the geometry estimator to reconstruct the entire world, \method uses geometry to {identify which pieces of history are relevant to each piece of the future}.
Our contributions are summarized as follows:
\begin{itemize}
\item We propose {\method}, a Geometry-Enabled Attention Routing framework that decouples visual memory from geometric addressing. By converting per-frame 3D priors into token-level cross-view correspondences, \method enables fine-grained access to relevant historical visual features without error-prone global 3D fusion.
\item We introduce {Geometric Correspondence Attention} for sparse token-level memory injection,
together with an Invisible Octree for visibility-aware filtering.
\item Extensive experiments demonstrate state-of-the-art visual quality, camera-control accuracy, and revisit consistency, enabling minute-long video generation along challenging user-specified trajectories with only lightweight adaptation of a pretrained video model.
\end{itemize}

\section{Related Work}
\paragraph{Implicit Camera-Controlled Video Generation.}
Early camera-controlled methods inject camera trajectories or user actions into video generation~\citep{he2024cameractrl,bai2025recammaster,yu2024gamefactory,li2025hunyuangamecrafthighdynamicinteractivegame}.
Subsequent streaming approaches extend generation to longer horizons~\citep{huang2025selfforcing,team2026advancing,he2025matrix}, but finite temporal context limits scene persistence.
To retain visual history, \citet{xiao2026worldmem,li2025vmem,yu2025cam,sun2025worldplay} retrieve historical frames as context, while \citet{hong2025relic,wu2026infiniteworld} compress history into compact latent.
These approaches preserve rich visual information but still require dense attention with an increasing set of historical tokens.

\vspace{-1.25 em}
\paragraph{Explicit Camera-Controlled Video Generation.}
Recent methods reconstruct historical observations into 3D memories and project them to target viewpoints as pixel-aligned conditions~\citep{wu2025video,zhao2026spatia,zhang2026worldstereo,lee20253dscenepromptingsceneconsistent,wang2026anchorweave,chen2026anyrecon}.
While providing explicit spatial grounding, these methods are vulnerable to accumulated reconstruction and fusion errors.
More closely related to our work, UCM~\citep{xu2026ucm} warps positional encodings to establish cross-view relationships, while Lyra 2.0~\citep{shen2026lyra} injects warped correspondence coordinates as token embeddings.
Our \method instead uses per-frame geometry to select visible historical tokens for sparse correspondence attention, retaining visual memory without global 3D fusion.
\section{Methods}

\subsection{Problem Formulation and Preliminaries}
\label{sec:preliminary}
\paragraph{Camera-Conditioned Autoregressive Video Generation.}
Given an initial frame $I_1$, text prompts $y$ and a long camera trajectory, our goal is to autoregressively synthesize visual observations with accurate camera control and long-term consistency with previously observed scene regions.

Following~\citet{wu2026framecrafter,chen2026anyrecon}, we encode each frame independently using the VAE~\citep{wan2025wan}, $z_i = E(I_i)$, which preserves frame-level alignment between visual observations and camera poses, particularly under large camera motions, enabling cross-view correspondences to be constructed directly at the latent-token level.

After generating $t$ frames, we maintain a streaming history bank with the historical latent observations and their camera parameters:
\begin{equation}
    \mathcal{B}_t = \{(z_i, c_i)\}_{i=1}^{t}.
\end{equation}

Given the next camera chunk $c_{t+1:t+k}$, a compact conditioning history $\mathcal{H}_t^{\mathrm{ret}}$ is retrieved from $\mathcal{B}_t$ according to its geometric coverage of the target views, as detailed in Sec.~\ref{sec:long_horizon}. The video generator then samples the next latent chunk as:
\begin{equation}
    z_{t+1:t+k}
    \sim
    p_\theta\!\left(
        \cdot
        \mid
        \mathcal{H}_t^{\mathrm{ret}},
        c_{t+1:t+k},
        y
    \right).
\end{equation}
The generated latents and their camera parameters are appended to $\mathcal{B}_t$, and the same procedure is repeated over successive chunks of the prescribed trajectory.

\begin{figure*}[t]
    \centering
    \scriptsize
    \includegraphics[width=0.975\textwidth]{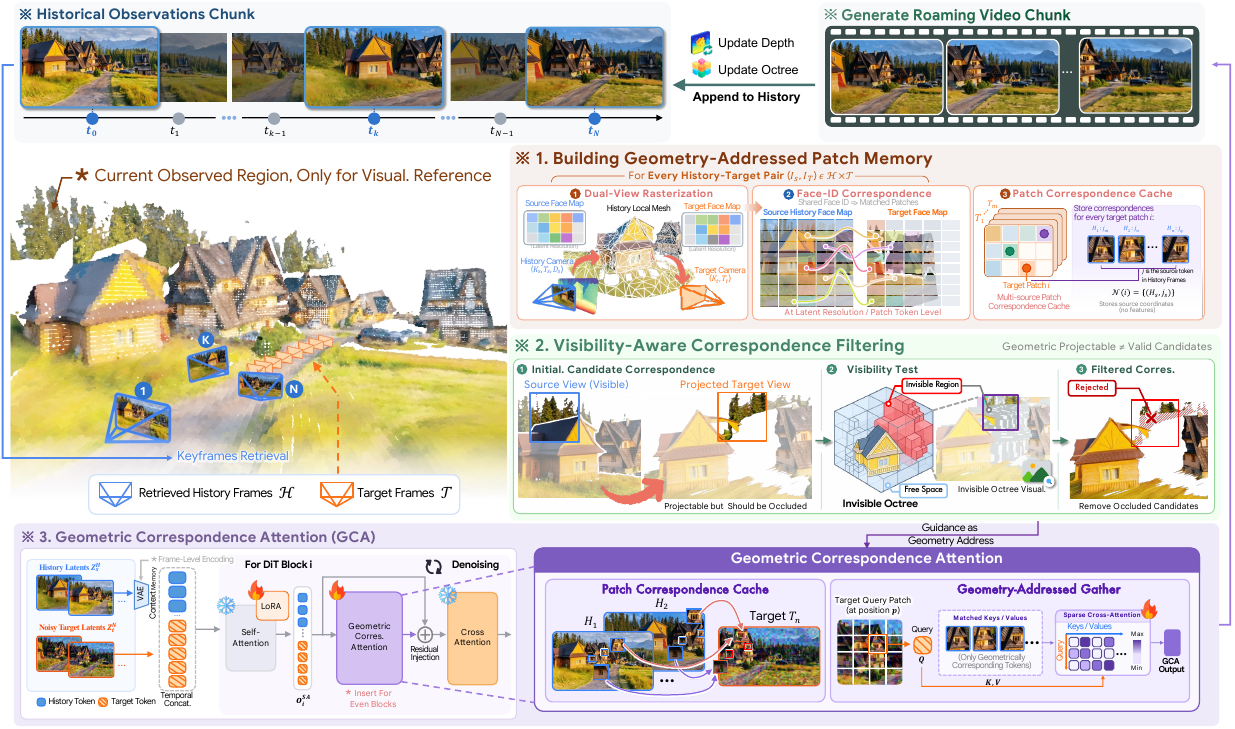}
    \caption{
        \textbf{System overview.}
        For each target chunk, \method constructs patch correspondences to retrieved history using per-frame geometry and filters occluded matches with the Invisible Octree. GCA then injects matched historical features into noisy target tokens during denoising, after which generated observations are appended to the history bank for continued rollout.
    }
    \label{fig:pipeline}
\end{figure*}

\vspace{-1.25 em}
\paragraph{Flow-Matching Objective.}
We train the video generator with the standard flow-matching objective~\citep{lipman2022flow}.
For a target latent block $\mathbf{z}_{t+1:t+k}$, we sample Gaussian noise
$\boldsymbol{\epsilon}\sim\mathcal{N}(\mathbf{0},\mathbf{I})$ and a flow timestep
$\tau\sim\mathcal{U}(0,1)$, and construct the linear interpolation:
\begin{equation}
    \mathbf{z}^{\tau}_{t+1:t+k}
    =
    (1-\tau)\mathbf{z}_{t+1:t+k}
    +
    \tau\boldsymbol{\epsilon}.
\end{equation}
The conditional velocity field is optimized as:
\begin{equation}
\label{eq:flow_matching}
\begin{aligned}
    \mathcal{L}_{\mathrm{FM}}
    =
    \mathbb{E}_{\mathbf{z},\boldsymbol{\epsilon},\tau}
    \Big[
        \big\|
            & v_{\theta}
            \left(
                \mathbf{z}^{\tau}_{t+1:t+k},
                \tau
                \mid
                \mathcal{H}_t,
                \mathbf{c}_{t+1:t+k}, y
            \right)
            \\
            & -
            \left(
                \boldsymbol{\epsilon}
                -
                \mathbf{z}_{t+1:t+k}
            \right)
        \big\|_2^2
    \Big].
\end{aligned}
\end{equation}

Although history retrieval bounds the number of conditioning frames, only a sparse and view-dependent subset of their tokens is relevant to each target region. The key problem is not merely which historical frames to retain, but how each target token should access the corresponding historical evidence. In the following, we introduce Geometry-Addressed Patch Memory to construct these fine-grained memory addresses from per-frame geometry.

\subsection{Geometry-Addressed Patch Memory}
\label{sec:patch_memory}

Given the retrieved history, our goal is to establish patch-level correspondences between historical and target views, which subsequently serve as explicit addresses for memory retrieval. Rather than constructing a globally fused scene representation, we derive these correspondences independently from the local geometry associated with each historical frame, as illustrated in Fig.~\ref{fig:pipeline}.

\vspace{-1.25 em}
\paragraph{Local Geometric Anchor.}
For each historical frame $s$, we associate its latent $z_s$ with an estimated depth map $D_s$, camera intrinsics $K_s$, and extrinsics $T_s$. We back-project $D_s$ into 3D and connect neighboring pixels according to the image-grid topology, producing a local triangular mesh $\mathcal{G}_s = (\mathcal{V}_s, \mathcal{F}_s)$, where $V_s$ and $F_s$ denote the mesh vertices and faces, respectively.
Since modern video VAEs aggressively compress the spatial resolution, e.g., by a factor of $16\times$, each latent token may cover pixels belonging to multiple surfaces.
We construct the mesh at the original image resolution to preserve geometric discontinuities that would otherwise be blurred by directly downsampling depth to the latent grid.

\vspace{-1.25 em}
\paragraph{Geometry-Guided Patch Correspondence Addressing.}
We then rasterize the same source mesh under both the source and target cameras at the latent spatial resolution $H_\ell \times W_\ell$, yielding two face-index maps:
\begin{equation}
\begin{aligned}
    F_s &=
    \mathcal{R}_{H_\ell \times W_\ell}
    (\mathcal{G}_s; K_s, T_s),\\
    F_{s\rightarrow t} &=
    \mathcal{R}_{H_\ell \times W_\ell}
    (\mathcal{G}_s; K_t, T_t),
\end{aligned}
\end{equation}
where each valid entry records the ID of the visible mesh face associated with a latent patch token.
Since both maps are rasterized from the same local geometry, shared face IDs naturally establish a token-level correspondence:
\begin{equation}
    \mathcal{C}_{s\rightarrow t}
    =
    \left\{
    (i,j)
    \;\middle|\;
    F_{s\rightarrow t}(i) = F_s(j)
    \right\}.
\end{equation}

Shared face identities thus provide a direct geometric bridge between source and target latent tokens, while preserving the fine spatial structure captured by the full-resolution source geometry.

\begin{figure*}[t]
    \centering
    \scriptsize
    \includegraphics[width=0.975\textwidth]{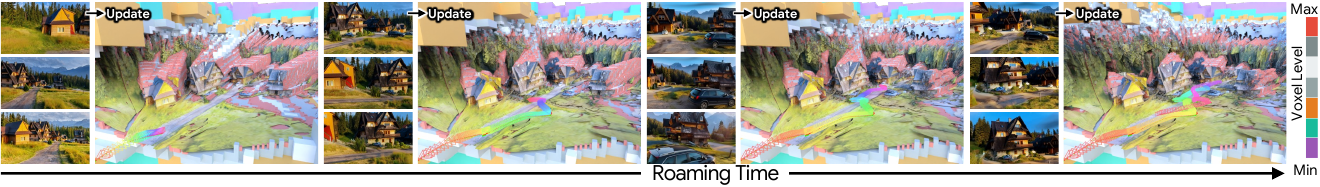}
    \caption{
        \textbf{Streaming update of the Invisible Octree.} Historical observations incrementally accumulate coarse visibility evidence along the exploration trajectory. For a target view, the resulting invisible-space boundary is used to reject projectable but occluded source-target correspondences.
    }
    \label{fig:long_horizon_octree}
\end{figure*}

\vspace{-1.25 em}
\paragraph{Multi-Source Patch Correspondence Cache.}
We construct $\mathcal{C}_{s \rightarrow t}$ for every selected history-target frame pair
and organize them into a patch correspondence cache:
\begin{equation}
    \mathcal{C}
    \in
    \mathbb{Z}^{N_t \times N_c \times H_\ell \times W_\ell \times 2},
\end{equation}
where $N_t$ and $N_c$ denote the numbers of target and retrieved historical condition frames.
For each target patch and historical frame, $\mathcal{C}$ stores the coordinate $(x_s,y_s)$ of its geometrically corresponding source patch, with unmatched entries marked as invalid.
Since the same scene region may have been observed from multiple historical viewpoints, a target patch can naturally admit multiple source correspondences.
We therefore retain all valid historical candidates.
Crucially, since all correspondences are derived independently from each source observation, geometric errors remain local instead of accumulating into persistent artifacts through global 3D fusion.

\subsection{Visibility-Aware Correspondence}
\label{sec:invisible_octree}

\begin{figure}[t]
    \centering
    \scriptsize
    \includegraphics[width=\linewidth]{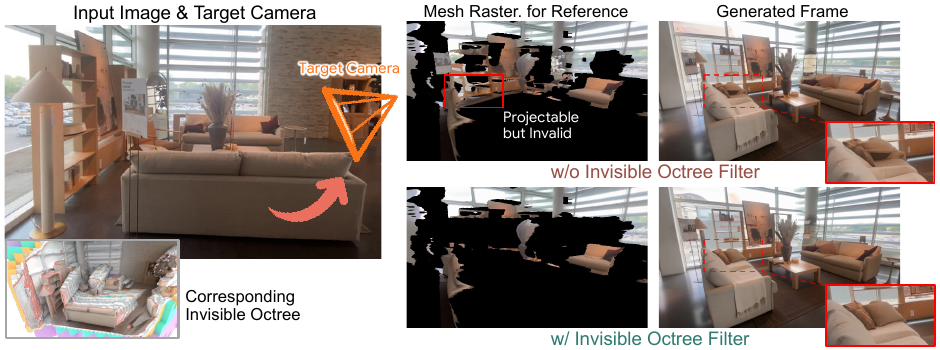}
    \caption{
        \textbf{Ablation of Invisible Octree.}
        The Invisible Octree removes projectable but occluded correspondences, preventing erroneous historical content from affecting target-view synthesis.
    }
    \label{fig:exp:ablation_octree}
\end{figure}

Geometric projectability does not necessarily imply target-view visibility.
As illustrated in Fig.~\ref{fig:pipeline},
a surface visible in source frame $s$ projects into target frame $t$ while being occluded by geometry unobserved in $s$.
Such candidates provide spatially incorrect historical evidence and should be removed before memory retrieval.

To validate these correspondences, we maintain an Invisible Octree as a sparse global visibility proxy. After each generated chunk, the estimated depth maps are used to incrementally update the octree with newly observed free-space and occlusion evidence, as illustrated in Fig.~\ref{fig:long_horizon_octree}.
The update is conservative: new observations only refine previously unknown or invisible regions, allowing the octree to expand with exploration without repeatedly overwriting established evidence. Its sparse and adaptive structure also supports long-horizon generation in unbounded environments.
\textbf{Please refer to Appendix for details on the construction and streaming update of the Invisible Octree.}

For a target camera, we use the Invisible Octree to determine the visibility mask under the target viewpoint.
Candidates lying behind the accumulated visibility boundary are rejected as occluded.
Importantly, the octree stores no appearance and never serves as a rendering condition; it only provides a coarse binary filter over correspondences.
\subsection{Geometric Correspondence Attention}
\label{sec:gca}

The correspondence cache above specifies \emph{where} each target patch should retrieve historical evidence from. Building on these memory addresses, we introduce \textbf{Geometric Correspondence Attention (GCA)} as an auxiliary memory pathway, which selectively aggregates only the geometrically matched historical tokens and injects them into target noisy tokens during denoising.

\vspace{-1.25 em}
\paragraph{Sparse Correspondence Attention.}
Following context-memory-based approaches~\citep{sun2025worldplay,yu2025cam}, we concatenate the retrieved historical frames (Sec.~\ref{sec:long_horizon}) as context latents with the noisy target latents along the temporal dimension.
Let $h_i^T$ denote the post-self-attention feature of target token $i$, and $h_j^H$ denote the corresponding feature of historical token $j$. From the patch correspondence cache, each target token obtains its geometrically matched historical candidates $\mathcal{N}(i)$. We then perform cross-attention:
\begin{equation}
q_i = W_Q h_i^T,
\qquad
k_j = W_K h_j^H,
\qquad
v_j = W_V h_j^H.
\end{equation}

The attention weight assigned to candidate $j$ is normalized only over the geometrically matched set, and the geometry-addressed memory feature is aggregated as:
\begin{equation}
\begin{aligned}
    \alpha_{ij}
    &=
    \frac{
        \exp\!\left(q_i^\top k_j / \sqrt{d}\right)
    }{
        \sum_{m \in \mathcal{N}(i)}
        \exp\!\left(q_i^\top k_m / \sqrt{d}\right)
    },
    \\
    o_i^{\mathrm{GCA}}
    &=
    \sum_{j \in \mathcal{N}(i)}
    \alpha_{ij} v_j.
\end{aligned}
\end{equation}

When multiple historical views observe the same target region, attention adaptively aggregates their complementary appearance information. If $\mathcal{N}(i)$ is empty, we set $o_i^{\mathrm{GCA}} = 0$, allowing the pretrained video model to synthesize unobserved content from its generative prior.

\vspace{-1.25 em}
\paragraph{Lightweight Residual Injection.}

\begin{figure*}[t]
    \centering
    \scriptsize
    \includegraphics[width=\textwidth]{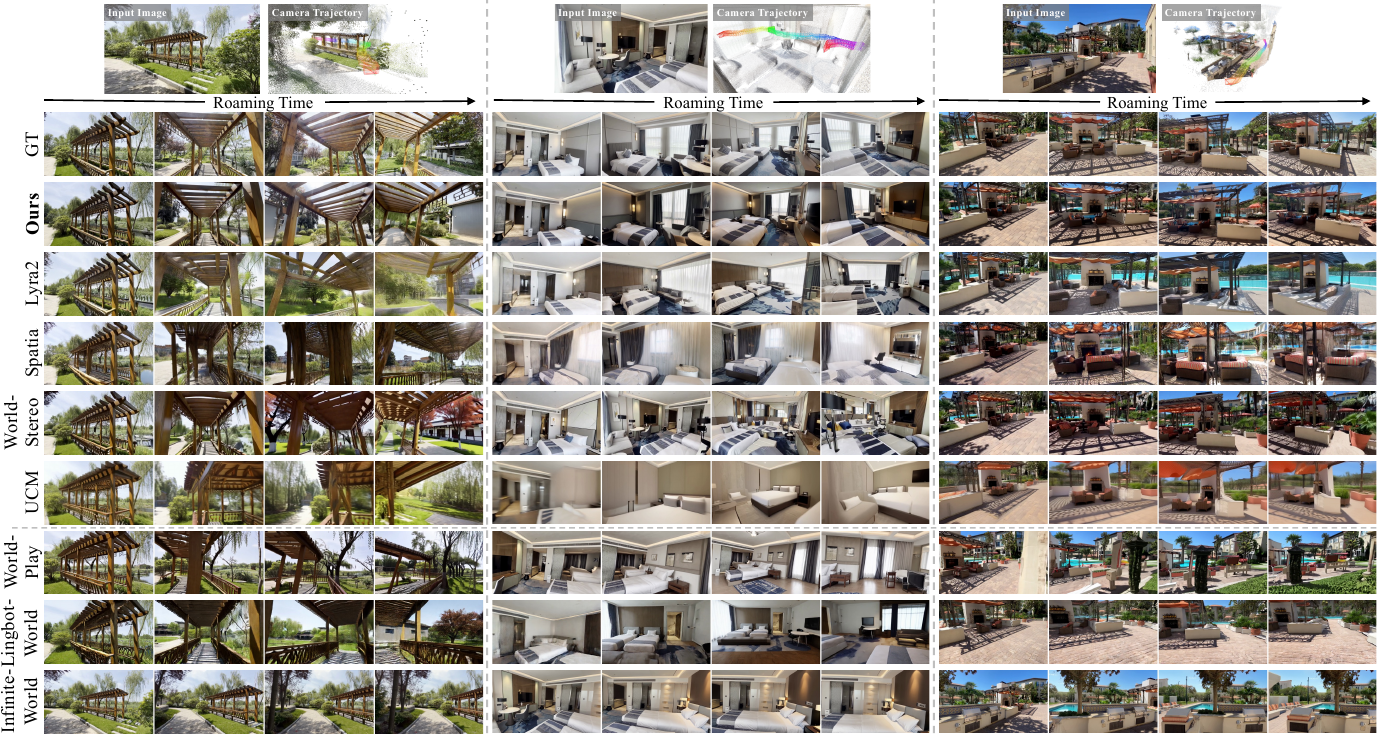}
    \caption{
        \textbf{Qualitative comparison on DL3DV-Eval.}
        Compared with explicit and implicit memory baselines, \method better preserves visual quality, camera adherence, and scene consistency throughout long-horizon generation.
        \textbf{See our project page for additional scenes and video comparisons.}
    }
    \vspace{1.5 em}
    \label{fig:exp:dl3dv_comparison}
\end{figure*}

\providecommand{\best}[1]{\textbf{#1}}
\providecommand{\sbest}[1]{\underline{#1}}

\begin{table*}[t]
    \centering
    \caption{
        \textbf{Quantitative comparison on DL3DV-Evaluation and WorldScore-Static.}
        \method performs favourably across the metrics.
        Best results are in \textbf{bold} and second are \underline{underlined}.
    }
    \label{tab:quantitative_comparison}

    \scriptsize
    \renewcommand{\arraystretch}{1.18}
    \setlength{\tabcolsep}{6.28pt}

    \begin{tabular*}{\textwidth}{@{\extracolsep{\fill}}lccc|ccc|cccc|cc@{}}
        \specialrule{.15em}{.1em}{.1em}

        &
        \multicolumn{6}{c|}{\textbf{DL3DV-Evaluation}}
        &
        \multicolumn{6}{c}{\textbf{WorldScore-Static}}
        \\

        \cmidrule(lr){2-7}
        \cmidrule(lr){8-13}

        \textbf{Method}
        & \textbf{SSIM} $\uparrow$
        & \textbf{LPIPS} $\downarrow$
        & \textbf{FVD} $\downarrow$
        & \textbf{TransErr} $\downarrow$
        & \textbf{RotErr} $\downarrow$
        & \textbf{ATE} $\downarrow$
        & \makecell[c]{\textbf{Content}\\\textbf{Align.} $\uparrow$}
        & \makecell[c]{\textbf{Photo.}\\\textbf{Cons.} $\uparrow$}
        & \makecell[c]{\textbf{Style}\\\textbf{Cons.} $\uparrow$}
        & \makecell[c]{\textbf{Subjective}\\\textbf{Quality} $\uparrow$}
        & \makecell[c]{\textbf{Revisit}\\\textbf{SSIM} $\uparrow$}
        & \makecell[c]{\textbf{Revisit}\\\textbf{LPIPS} $\downarrow$}
        \\

        \midrule
        Lyra2
        & 0.3359 & \sbest{0.5097} & 975.89
        & \sbest{0.0157} & 0.1721 & 0.2514
        & 0.6319 & 0.9357 & \sbest{0.8600}
        & \sbest{0.5017} & 0.3941 & 0.3218
        \\
        Spatia
        & 0.3081 & 0.5422 & 1074.13
        & 0.0617 & 0.6973 & 1.1204
        & 0.6331 & 0.8588 & \sbest{0.8600}
        & 0.5012 & 0.4407 & 0.3541
        \\
        WorldStereo
        & 0.3061 & 0.5502 & \sbest{846.46}
        & 0.0239 & \sbest{0.1717} & \sbest{0.2212}
        & \sbest{0.7003} & 0.0837 & 0.8300
        & 0.5013 & \sbest{0.6253} & \sbest{0.2193}
        \\
        UCM
        & \sbest{0.3412} & 0.6007 & 1431.67
        & 0.0377 & 0.5184 & 0.6410
        & 0.6773 & 0.9692 & 0.7300
        & 0.5005 & 0.3412 & 0.4890
        \\

        \midrule
        HY-WorldPlay
        & 0.2452 & 0.6213 & 1385.23
        & 0.0508 & 0.6854 & 0.9634
        & 0.5104 & 0.7939 & 0.1900
        & \best{0.5018} & 0.2433 & 0.7288
        \\
        Lingbot-World
        & 0.2608 & 0.6168 & 1309.27
        & 0.0417 & 0.5988 & 0.7449
        & 0.5564 & 0.0889 & 0.6800
        & 0.5006 & 0.2495 & 0.7677
        \\
        Infinite-World
        & 0.2532 & 0.6399 & 1582.43
        & 0.1072 & 0.8473 & 2.3744
        & 0.6302 & \sbest{0.9728} & 0.7000
        & 0.5011 & 0.2536 & 0.7098
        \\

        \midrule
        \best{\method}
        & \best{0.3645}
        & \best{0.4459}
        & \best{837.59}
        & \best{0.0116}
        & \best{0.1228}
        & \best{0.0436}
        & \best{0.7423}
        & \best{0.9732}
        & \best{0.8700}
        & \best{0.5018}
        & \best{0.6489}
        & \best{0.2019}
        \\

        \specialrule{.15em}{.1em}{.1em}
    \end{tabular*}
\end{table*}

GCA is inserted after the original self-attention at selected DiT blocks, as shown in Fig.~\ref{fig:pipeline}. Its output is projected back to the backbone feature space and injected through a residual connection:
\begin{equation}
    \widetilde{h}_i^T
    =
    h_i^T
    +
    W_O o_i^{\mathrm{GCA}},
\end{equation}
where $W_O$ projects the feature back to the DiT feature space.
The backbone self-attention retains its pretrained spatiotemporal modeling,
while GCA supplies a sparse geometry-addressed correction that anchors target features to relevant historical evidence.

\subsection{Robust Training with Degraded History}
\label{sec:degraded_history}

During training, target chunks are conditioned on ground-truth history, whereas autoregressive inference relies on previously generated observations. This train–inference discrepancy causes generation errors to enter the history memory and progressively accumulate over long rollouts.

To expose the model to imperfect yet semantically consistent history, we introduce degraded-history augmentation.
With probability $p_{\mathrm{deg}}$, we corrupt the historical latents $\mathbf{z}_{\mathcal{H}}$ with a randomly sampled low noise level $\tau_h \sim \mathcal{U}(0, \tau_{\mathrm{max}})$:
\begin{equation}
    \mathbf{z}^{\tau_h}_{\mathcal{H}}
    =
    (1-\tau_h)\mathbf{z}_{\mathcal{H}}
    +
    \tau_h \boldsymbol{\epsilon}_{\mathcal{H}},
    \qquad
    \boldsymbol{\epsilon}_{\mathcal{H}} \sim \mathcal{N}(0,\mathbf{I}).
\end{equation}
We then apply one reverse-flow step with the current model to obtain the degraded history:
\begin{equation}
    \hat{\mathbf{z}}_{\mathcal{H}}
    =
    \mathrm{sg}[
    \mathbf{z}^{\tau_h}_{\mathcal{H}}
    -
    \tau_h\,
    v_{\theta}
    \left(
        \mathbf{z}^{\tau_h}_{\mathcal{H}},
        \tau_h
    \right)
    ]
    .
\end{equation}
The resulting $\hat{\mathbf{z}}_{\mathcal{H}}$ replaces $\mathbf{z}_{\mathcal{H}}$ as the conditioning context, while keeping the flow-matching objective for the target chunk unchanged.
This exposes the model to the mild distortions encountered during rollout and improves robustness to accumulated history errors in long-horizon inference.

\begin{figure*}[t]
    \centering
    \scriptsize
    \vspace{-1.5 em}
    \includegraphics[width=0.975\textwidth]{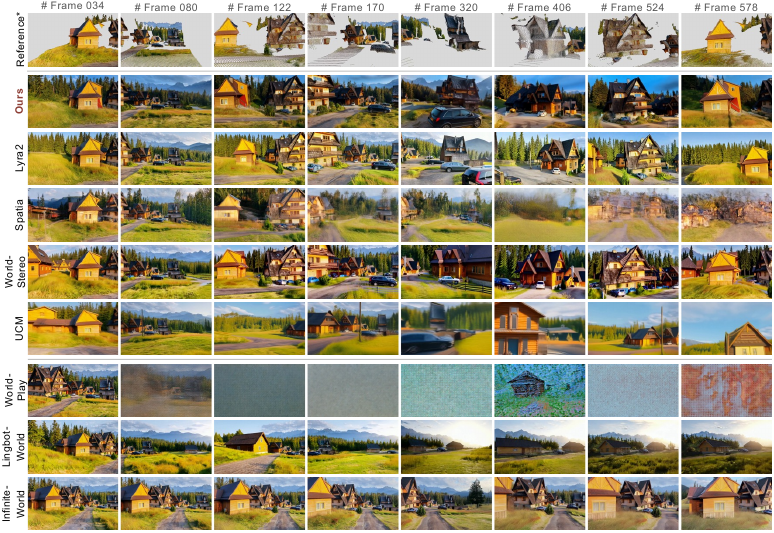}
    \caption{
        \textbf{Out-of-domain qualitative comparison for minute-long generation.}
        \method maintains visual fidelity and scene consistency over extended camera trajectories, while competing methods exhibit progressive drift and visual degradation.
        *First-frame geometry projected along the target trajectory for viewpoint reference only.
        \textbf{See our project page for additional video comparisons.}
    }

    \label{fig:exp:long_horizon_comparison}
\end{figure*}

\subsection{Long-Horizon Inference}
\label{sec:long_horizon}

For long-horizon inference, we adopt a streaming strategy that bounds the historical context through keyframe retrieval and continuously updates the memory.

\vspace{-1.25 em}
\paragraph{Keyframe History Retrieval.}
As the rollout progresses, retaining all historical frames introduces increasing computational cost and substantial view redundancy. We therefore retrieve a compact history according to its geometric coverage of the upcoming target chunk.
Specifically, we project each historical frame's local geometry onto the target views, and greedily select frames that maximize newly covered regions.
Meanwhile, the initial frame and the latest frame are retained to preserve scene identity and inter-chunk continuity.
The retrieved history frames are thus
$
\mathcal{H}^{\mathrm{ret}}_t
=
\big[
\mathcal{H}_{\mathrm{first}},
\mathcal{H}_{\mathrm{coverage}},
\mathcal{H}_{\mathrm{last}}
\big].
$

\vspace{-1.25 em}
\paragraph{Streaming Memory Update.}
After each chunk is generated, new frames are appended to history bank with camera parameters and depths estimated by a frozen 3D foundation model~\citep{depthanything3}.
Invisible Octree is simultaneously updated with newly observed geometry.
This retrieve–generate–update procedure is repeated for subsequent chunks, enabling long-horizon autoregressive generation.

\section{Experiments}

\subsection{Experimental Setup}
\label{sec:exp_setup}

\paragraph{Dataset.}
We train \method on DL3DV-10K~\citep{ling2024dl3dv}, a large-scale real-world dataset with diverse camera trajectories. Each sequence is divided into 55-frame clips at a resolution of $480\times832$. We employ Depth Anything 3~\citep{depthanything3} to estimate camera poses and per-frame depths, and generate video captions with Qwen3-VL-8B-Instruct~\citep{bai2025qwen3}.

We construct training samples under two conditioning modes. In image-to-video (I2V) mode, we train on the first 32 frames, with the initial frame providing the geometry for initializing the Invisible Octree and establishing correspondences with the target views. In history-to-video (H2V) mode, the first 32 frames constitute the history bank, from which nine keyframes are retrieved following Sec.~\ref{sec:long_horizon}; the remaining 23 frames serve as generation targets. The retrieved keyframes are used to construct the multi-source patch correspondence cache, while the complete history bank is used to build the global Invisible Octree for visibility-aware correspondence filtering.

\begin{figure*}[t]
    \centering
    \scriptsize
    \vspace{-1.5 em}
    \includegraphics[width=\textwidth]{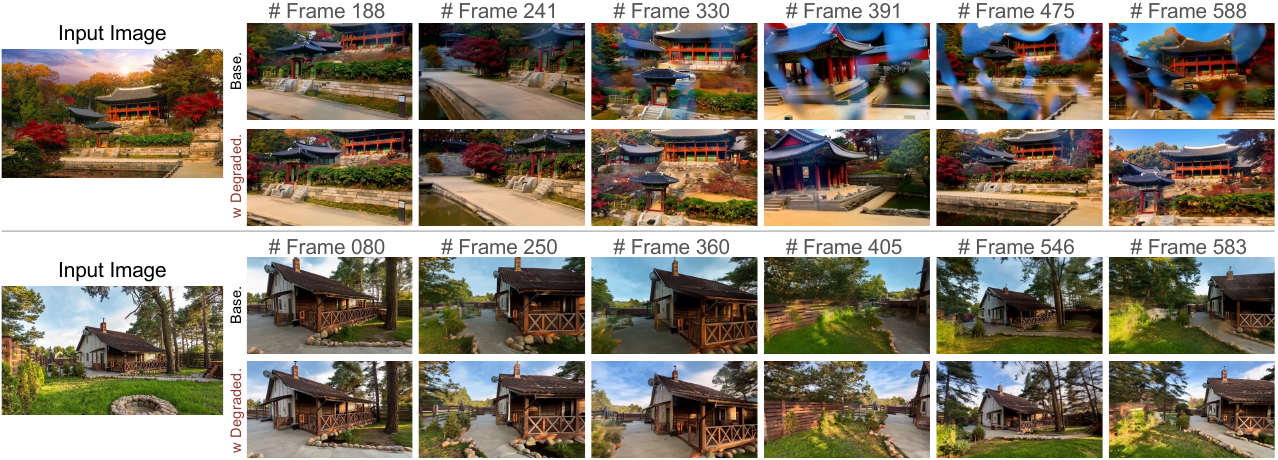}
    \caption{
        \textbf{Ablation of Degraded-History Augmentation.}
        Training with degraded history mitigates error accumulation and improves visual stability during long-horizon autoregressive generation.
    }
    \label{fig:exp:ablation_degraded}
\end{figure*}

\vspace{-1.25 em}
\paragraph{Implementation Details.}
We adopt Wan2.1-I2V-14B~\citep{wan2025wan} as the pretrained backbone and keep all original parameters frozen.
To adapt the backbone to frame-aligned VAE latents, we introduce rank-32 LoRA adapters~\citep{hu2021lora}, and we insert GCA modules with a hidden dimension of 640 into every even-indexed DiT block (262M parameters for GCA, only 1.9\% of the backbone).
The LoRA adapters and GCA modules are jointly optimized for 10K iterations.

During training, I2V and H2V samples are drawn with probabilities of $30\%$ and $70\%$, respectively.
Starting from iteration 8K, degraded-history augmentation is further applied to H2V samples with
a probability $p_{\mathrm{deg}}=40\%$ and $\tau_{\mathrm{max}}=0.3$.
We optimize the model using AdamW on 32 GPUs with a learning rate of $1\times10^{-4}$ and a linear warm-up over the first 1K iterations.
\subsection{Quantitative Evaluation}

\label{sec:exp_comparison}
\paragraph{Baselines and Metrics.}
We compare \method with recent camera-controlled long-horizon video generation methods equipped with memory mechanisms.
Explicit 3D baselines include Lyra2~\citep{shen2026lyra}, Spatia~\citep{zhao2026spatia}, HY-WorldStereo~\citep{zhang2026worldstereo}, and UCM~\citep{xu2026ucm},
while implicit baselines include HY-WorldPlay~\citep{sun2025worldplay}, Lingbot-World~\citep{team2026advancing}, and Infinite-World~\citep{wu2026infiniteworld}.

We first evaluate all methods on DL3DV-Evaluation~\citep{ling2024dl3dv}. Given the same initial frame, each method autoregressively generates subsequent frames along the ground-truth camera trajectory.
We report SSIM~\citep{wang2004image}, LPIPS~\citep{zhang2018unreasonable}, and FVD~\citep{unterthiner2018towards} to evaluate frame-level fidelity and temporal quality.
For camera-control accuracy, we recover camera poses from the generated videos using ViPE~\citep{huang2025vipe} and align them with the ground-truth trajectories using the Umeyama transformation~\citep{umeyama1991least}. We then report TransErr, RotErr, and Absolute Trajectory Error (ATE).

We further evaluate long-horizon revisitation on 50 randomly sampled scenes from the WorldScore Static set~\citep{duan2025worldscore} using closed-loop camera trajectories. We report Content Alignment, Subjective Quality, Style Consistency, and Photometric Consistency, together with Revisit SSIM and Revisit LPIPS, which are computed between the generated revisit frame and the reference observation at the matched camera pose to assess the recovery of previously observed scene content.

\begin{wrapfigure}{r}{0.4\linewidth}
    \centering
    \scriptsize
    \vspace{-1.5 em}
    \includegraphics[width=\linewidth]{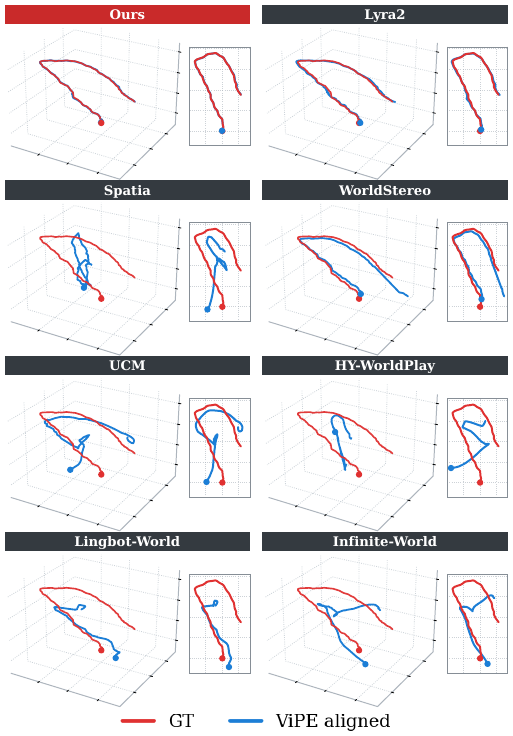}
    \caption{
        \textbf{Camera alignment.} \method achieves the best camera alignment on DL3DV-Evaluation.
    }
    \label{fig:exp:camera_align}
\end{wrapfigure}

\vspace{-1.25 em}
\paragraph{Quantitative and Qualitative Comparison.}
As shown in Tab.~\ref{tab:quantitative_comparison}, \method performs favorably across the evaluated metrics.
Our method reduces ATE by 80.3\% compared with
strongest baseline, demonstrating improved trajectory adherence, as further illustrated in Fig.~\ref{fig:exp:camera_align}.
\method achieves the best content alignment, photometric consistency, style consistency, and revisit performance on WorldScore, which demonstrates \method faithfully
recovers previously observed content after long temporal gaps without compromising overall generation quality.

The qualitative comparisons in Fig.~\ref{fig:exp:dl3dv_comparison} and \ref{fig:exp:long_horizon_comparison} further demonstrate the advantages of \method over existing methods.
Lyra2 provides competitive camera control but develops visual degradation and increasing trajectory drift under rapid or extended camera motion, while Spatia and WorldStereo exhibit progressive scene distortions, consistent with their vulnerability to accumulated reconstruction errors.
UCM initially preserves coherent appearance but gradually deviates over extended rollouts.
HY-WorldPlay and Lingbot-World struggle with both camera control and revisitation, whereas Infinite-World maintains comparatively stable appearance but insufficiently follows the precise trajectory.
In particular, WorldStereo and Lingbot-World exhibit abrupt appearance changes and pronounced flicker during camera rotations, consistent with their low Photometric Consistency scores.
These limitations become more pronounced during minute-long explorations with rapid camera motion (Fig.~\ref{fig:exp:long_horizon_comparison}): most baselines exhibit severe visual degradation and develop increasing camera drift. \method maintains coherent appearance and accurate camera control, supporting the effectiveness of separating visual memory from geometric addressing for long-horizon generation.
\textbf{\textcolor{red}{Please refer to our project page} for additional scenes and more extensive video comparisons.}
\begin{table}[t]
    \centering
    \vspace{2.0 em}
    \caption{
        \textbf{Ablation study on DL3DV-Evaluation.}
        We ablate the contribution of each key component of \method.
    }
    \label{tab:ablation}
    \small
\renewcommand{\arraystretch}{1.15}
\setlength{\tabcolsep}{3.2pt}

\begin{adjustbox}{max width=\linewidth}
\begin{tabular}{@{}lccc|ccc@{}}
    \specialrule{.15em}{.1em}{.1em}

    \textbf{Variant}
    & \textbf{SSIM} $\uparrow$
    & \textbf{LPIPS} $\downarrow$
    & \textbf{FVD} $\downarrow$
    & \textbf{TransErr} $\downarrow$
    & \textbf{RotErr} $\downarrow$
    & \textbf{ATE} $\downarrow$
    \\

    \midrule

    w/o GCA
    & 0.1324 & 0.6549 & 1201.32
    & 0.0406 & 0.4732 & 0.3471
    \\

    w/ Dense Attention
    & 0.2749 & 0.5995 & 913.64
    & 0.0351 & 0.4347 & 0.2924
    \\

    w/o Invisible Octree
    & 0.3350 & 0.4869 & 878.28
    & 0.0153 & 0.1551 & 0.0729
    \\

    w/o Degraded History
    & 0.3459 & 0.4763 & 1165.99
    & 0.0127 & 0.1492 & 0.0664
    \\

    w/ Corres. Disturbance
    & 0.3597 & 0.4513 & 842.15
    & 0.0121 & 0.1273 & 0.0464
    \\

    \midrule

    \textbf{\method}
    & \textbf{0.3645}
    & \textbf{0.4459}
    & \textbf{837.59}
    & \textbf{0.0116}
    & \textbf{0.1228}
    & \textbf{0.0436}
    \\

    \specialrule{.15em}{.1em}{.1em}
\end{tabular}
\end{adjustbox}

\end{table}

\begin{figure}[t]
    \vspace{0pt}
    \centering
    \includegraphics[width=\linewidth]{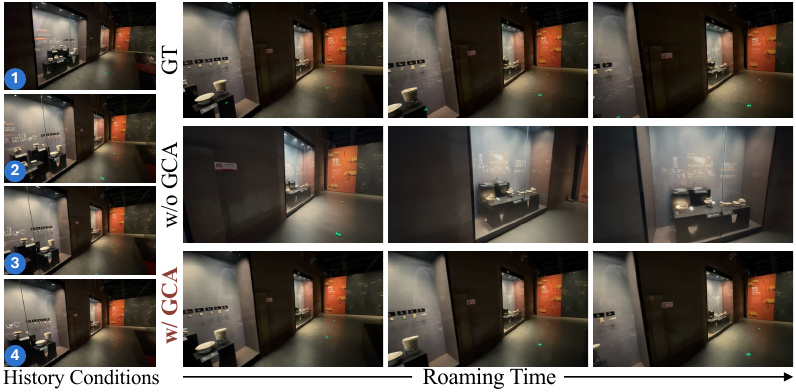}
    \caption{
        \textbf{Ablation of GCA.}
        GCA improves temporal stability and adherence to the prescribed camera trajectory.
    }
    \label{fig:exp:ablation_gca}
\end{figure}

\subsection{Ablation Study}
\label{sec:exp_ablation}

We evaluate the contribution of each key component of \method:
\vspace{-1.25 em}
\paragraph{Without Geometric Correspondence Attention (GCA).}
We train an ablated variant without GCA while retaining the backbone's dense attention over the concatenated history and target tokens. As shown in Fig.~\ref{fig:exp:ablation_gca}, disabling GCA's residual injection increases temporal instability and deviation from the camera motion.
The quantitative degradation in
Tab.~\ref{tab:ablation} further confirms the contribution of GCA to camera-control accuracy and long-horizon consistency.

\vspace{-1.25 em}
\paragraph{Variant with Dense History Attention.} To isolate geometric addressing from the additional capacity of GCA, we train a variant that retains the same GCA module and parameter count but replaces correspondence-restricted attention with dense attention over full retrieved historical context tokens.
The resulting degradation in camera control and visual fidelity (Tab.~\ref{tab:ablation}) confirms the importance of our sparse, geometry-addressed memory retrieval.

\vspace{-1.25 em}
\paragraph{Without Invisible Octree.}
Under large viewpoint changes as shown in Fig.~\ref{fig:exp:ablation_octree}, the projectable yet occluded
correspondences not only introduce local appearance errors but also propagate structural inconsistencies through the autoregressive history. The Invisible Octree suppresses this error propagation by rejecting candidates that conflict with accumulated visibility evidence.

\begin{figure}[t]
    \centering
    \scriptsize
    \includegraphics[width=\linewidth]{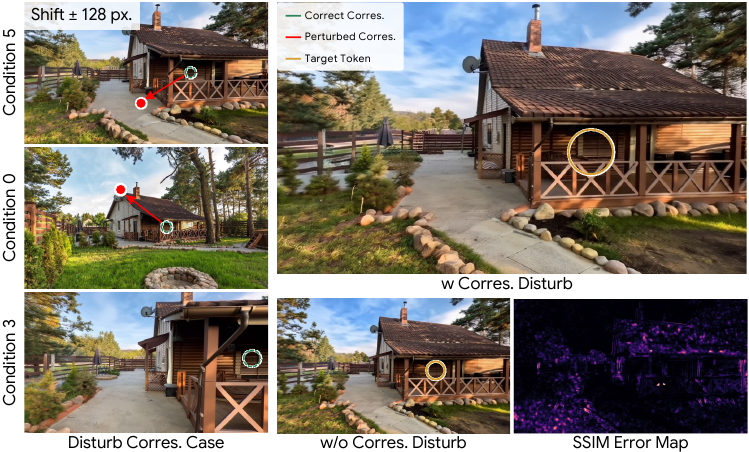}
    \caption{
        \textbf{Ablation of Correspondence Disturbance.}
        We perturb candidates in multi-observed target regions by 64--128 pixels while retaining only 2--3 valid matches out of 9 conditions.
        The figure visualizes one such perturbation and the resulting error map; GCA recovers the target content with minimal deviation from the undisturbed result.
    }
    \label{fig:exp:ablation_disturb}
\end{figure}

\vspace{-1.25 em}
\paragraph{Without Degraded-History Augmentation.}
As shown in Fig.~\ref{fig:exp:ablation_degraded}, training exclusively with ground-truth history leads to droplet-like floaters and visual degradation during long-horizon rollout. Degraded-history augmentation mitigates these artifacts by exposing the model to imperfect historical observations, substantially improving robustness and visual stability over extended generation.

\vspace{-1.25 em}
\paragraph{Robustness to Correspondence Disturbance.}
Depth errors accumulated during autoregressive rollout may introduce inaccurate history-to-target correspondences.
For target regions observed in multiple historical frames,
we retain only 2--3 valid observations out of 9 and perturb the rest by 64--128 pixels to simulate such errors.
Despite these strong perturbations, \method maintains comparable performance as shown in Fig.~\ref{fig:exp:ablation_disturb} and Tab.~\ref{tab:ablation}, indicating that the learned attention of GCA suppresses feature-inconsistent distractors and preserves valid historical evidence.

\section{Conclusion}

We introduced \method, a geometry-enabled attention routing framework for long-horizon camera-controlled video generation. \method uses per-frame geometry to establish local history-to-target token correspondences, and leverages lightweight Geometric Correspondence Attention to retrieve and integrate relevant historical features during denoising. By treating geometry as an address rather than persistent memory, \method avoids error accumulation from global 3D fusion while enabling minute-long generation with precise camera control and consistent scene persistence. \method currently relies on an external 3D model for depth estimation, introducing additional computational overhead. A promising direction for future work is to jointly learn geometry and memory addressing within the generative model for more robust long-horizon generation.

{
    \small
    \bibliographystyle{ieeenat_fullname}
    \bibliography{main}
}

\appendix
\clearpage
\maketitlesupplementary

This appendix provides additional details and results for GEAR. Appendix \ref{sec:appendix:additional_method_details} describes the complete streaming generation procedure (\ref{sec:appendix:additional_method_details:end_to_end_streaming}), including frame-aligned encoding (\ref{sec:appendix:additional_method_details:frame_aligned_vae}), correspondence construction (\ref{sec:appendix:additional_method_details:correspond_build_details}), visibility filtering (\ref{sec:appendix:additional_method_details:invisible_octree}), keyframe retrieval (\ref{sec:appendix:additional_method_details:keyframe_retrieval}), and depth updates (\ref{sec:appendix:additional_method_details:depth_update}). Appendix \ref{sec:appendix:dataset_details} reports dataset preprocessing, training settings, and inference costs. Appendix \ref{sec:appendix:evaluation_protocols} details the evaluation protocols and baseline configurations. Appendix \ref{sec:appendix:more_results} presents additional qualitative comparisons and long-horizon generation results.
\textbf{Additional video results and visualizations of our method details are available on our project page: \projectpage.}

\begin{algorithm*}[t]
    \caption{End-to-end streaming generation with \method}
    \label{alg:streaming_generation}
    \begin{algorithmic}[1]
    \Require History bank $\mathcal{B}_t=\{(I_s,z_s,c_s,D_s)\}_{s=1}^{t}$;
    future cameras $c_{t+1:T}$; text condition $y$;
    Invisible Octree $\mathcal{O}_t$; chunk length $k$
    \Ensure Generated frames $\hat I_{t+1:T}$ and updated history bank
    $\mathcal{B}_T$

    \While{$t<T$}
        \State $\mathcal{T}\gets\{t+1,\ldots,\min(t+k,T)\}$
        \State $\mathcal{H}\gets
            \Call{RetrieveKeyframes}{\mathcal{B}_t,\{c_u\}_{u\in\mathcal{T}}}$
            \Comment{Greedy target-view coverage; retain first and latest frames}

        \ForAll{$s\in\mathcal{H}$}
            \State $G_s\gets\Call{BuildLocalMesh}{D_s,c_s}$
                \Comment{Back-project depth and connect neighboring pixels}
            \State $F_s\gets
                \Call{RasterizeFaceIDs}{G_s,c_s,H_\ell,W_\ell}$
        \EndFor

        \State Initialize $\mathcal{C}[u,s,i]\gets\bot$ for
            $u\in\mathcal{T}$, $s\in\mathcal{H}$, and target patches $i$
        \ForAll{$u\in\mathcal{T}$}
            \State $M_u^{\mathrm{inv}}\gets
                \Call{ProjectInvisibleOctree}{\mathcal{O}_t,c_u}$
                \Comment{Target-view invisible mask}
            \ForAll{$s\in\mathcal{H}$}
                \State $F_{s\rightarrow u}\gets
                    \Call{RasterizeFaceIDs}{G_s,c_u,H_\ell,W_\ell}$
                \State $\mathcal{C}_{s\rightarrow u}\gets
                    \Call{MatchFaceIDs}{F_s,F_{s\rightarrow u}}$
                    \Comment{Store matched source-patch coordinates}
                \State $\mathcal{C}_{s\rightarrow u}\gets
                    \Call{FilterOccluded}{\mathcal{C}_{s\rightarrow u},
                    M_u^{\mathrm{inv}}}$
                \State Write valid matches from
                    $\mathcal{C}_{s\rightarrow u}$ into $\mathcal{C}[u,s,:]$
            \EndFor
        \EndFor

        \State $z_{\mathcal H}\gets\{z_s\}_{s\in\mathcal H}$;
            $x^{(0)}\sim\mathcal{N}(0,I)$
        \State Choose a denoising schedule
            $1=\tau_0>\tau_1>\cdots>\tau_{25}=0$
        \For{$n=0,\ldots,24$}
            \State $v^{(n)}\gets
                v_\theta(x^{(n)},\tau_n
                \mid z_{\mathcal H},\{c_u\}_{u\in\mathcal T},y,\mathcal C)$
                \Comment{GCA uses the filtered cache}
            \State $x^{(n+1)}\gets
                \Call{SolverStep}{x^{(n)},v^{(n)},\tau_n,\tau_{n+1}}$
        \EndFor

        \State $\hat z_{\mathcal T}\gets x^{(25)}$;
            $\hat I_{\mathcal T}\gets\Call{DecodeFramewise}{\hat z_{\mathcal T}}$
        \State $\hat D_{\mathcal T}\gets
            \Call{DepthAnything3}{\hat I_{\mathcal T}}$
        \ForAll{$u\in\mathcal T$}
            \State $\mathcal{B}_u\gets
                \mathcal{B}_{u-1}\cup
                \{(\hat I_u,\hat z_u,c_u,\hat D_u)\}$
        \EndFor
        \State $\mathcal{O}_{\max\mathcal T}\gets
            \Call{UpdateInvisibleOctree}{
            \mathcal{O}_t,\{(\hat D_u,c_u)\}_{u\in\mathcal T}}$
        \State $t\gets\max\mathcal T$
    \EndWhile
    \State \Return $\hat I_{t_0+1:T},\mathcal{B}_T$
    \end{algorithmic}
\end{algorithm*}

\section{Additional Method Details}
\label{sec:appendix:additional_method_details}

\subsection{End-to-End Streaming Generation}
\label{sec:appendix:additional_method_details:end_to_end_streaming}

We summarize the complete inference procedure in
Algorithm~\ref{alg:streaming_generation}. At generation step $t$, the
history bank contains the observed frames, their frame-aligned VAE
latents, camera parameters, and estimated depths:
$\mathcal{B}_t=\{(I_s,z_s,c_s,D_s)\}_{s=1}^{t}$, where
$c_s=(K_s,T_s)$ denotes the camera intrinsics and pose. The Invisible
Octree $\mathcal{O}_t$ accumulates visibility evidence from the
\emph{complete} history bank.

For each upcoming camera chunk, we retrieve a compact set of historical
frames according to their geometric coverage of the target views, while
retaining the first and latest frames for scene identity and inter-chunk
continuity. Each retrieved frame independently provides a local mesh
constructed from its depth map. Rasterizing this mesh under the source
and target cameras yields face-ID maps, from which we build a
multi-source patch correspondence cache. We then project the Invisible
Octree into each target view and remove candidates marked as occluded
by the resulting visibility mask.
The retrieved history latents, noisy target latents, and filtered correspondence cache jointly condition the DiT model throughout $N_{\mathrm{denoise}}=25$ denoising steps.
Finally, we decode the generated latents, estimate their depths
with Depth Anything 3~\citep{depthanything3}, append the new observations to the history bank,
and update the Invisible Octree before processing the next chunk.
\subsection{Frame-Aligned VAE Encoding}
\label{sec:appendix:additional_method_details:frame_aligned_vae}

GCA requires the geometric correspondences constructed in
Sec.~\ref{sec:patch_memory} to address latent tokens associated with
specific camera views. The original Wan VAE~\citep{wan2025wan} encodes
the first video frame separately but temporally compresses subsequent
frames by a factor of 4. Consequently, a latent frame after the
first generally aggregates observations captured at different
camera poses. Assigning a single pose to that latent frame cannot
provide an exact geometric interpretation for all observations.
The ambiguity becomes more pronounced under rapid camera motion, when
the aggregated frames may depict substantially different scene regions.
Thus, correspondences computed from frame-level depth and camera
parameters cannot be unambiguously transferred to temporally
compressed latent tokens.

Following~\citep{wu2026framecrafter,chen2026anyrecon}, we instead apply
the pretrained Wan VAE's single-frame encoding path independently to
every video frame. Each frame is treated as a separate one-frame input,
bypassing temporal compression while retaining the VAE's spatial
encoding:
\begin{equation}
    z_f = \mathcal{E}_{\mathrm{Wan}}(I_f), \qquad f=1,\ldots,F,
\end{equation}
where $\mathcal{E}_{\mathrm{Wan}}$ denotes the VAE applied to an
individual frame. This produces $F$ latent frames for $F$ video frames,
so each latent frame has a unique associated image, depth map, and
camera pose. Within that frame, each spatial latent token also has a
well-defined location on the image grid. We can therefore rasterize
per-frame geometry at the latent resolution and use the resulting
source--target patch correspondences to index historical tokens
directly during GCA. The generated latents are likewise decoded
frame by frame to preserve the same alignment at inference.
\subsection{Geometry-Addressed Correspondence Construction}
\label{sec:appendix:additional_method_details:correspond_build_details}

\paragraph{Depth back-projection and mesh connectivity.}
For a historical frame $s$, let $D_s$ be its estimated depth map and
$c_s=(K_s,T_s)$ its camera parameters. We construct a separate local
mesh $G_s=(V_s,F_s)$ from this observation. For each pixel
$p=(u,v)$ with a finite, positive depth, we back-project its pixel
center into world coordinates:
\begin{equation}
    \mathbf{x}_s(p)
    = T_s^{-1}
      \left(D_s(p)K_s^{-1}
      \begin{bmatrix}u+\tfrac{1}{2}\\v+\tfrac{1}{2}\\1\end{bmatrix}\right),
    \label{eq:supp_backprojection}
\end{equation}
where $T_s$ denotes the world-to-camera transform and homogeneous
coordinates are understood. Each valid pixel contributes one vertex.
We then split each $2\times2$ image-grid cell along a fixed diagonal
to form two candidate triangles. This retains the spatial resolution
of the depth map during mesh construction rather than smoothing depth
discontinuities by first downsampling into the latent grid.

\vspace{-1.25 em}
\paragraph{Invalid-depth and discontinuity filtering.}
A candidate triangle is discarded if any of its vertices has an
invalid depth. We also remove triangles that would connect surfaces
across a sharp depth discontinuity. Specifically, for a triangle
$f$ with pixel vertices $p_1,p_2,p_3$, we retain it only if
\begin{equation}
    \max_{(p_i,p_j)\in E(f)}
    \frac{|D_s(p_i)-D_s(p_j)|}
         {\min\!\left(D_s(p_i),D_s(p_j)\right)}
    \leq \tau_{\mathrm{disc}},
    \label{eq:supp_depth_discontinuity}
\end{equation}
where $E(f)$ contains its three edges and
$\tau_{\mathrm{disc}}$ is the relative depth-discontinuity threshold.
This filtering prevents triangles from spanning foreground--background
boundaries and producing correspondences through unsupported geometry.
Meshes are constructed independently for each historical frame; their
vertices and faces are not fused across observations.

\begin{figure*}[t]
    \centering
    \scriptsize
    \includegraphics[width=\textwidth]{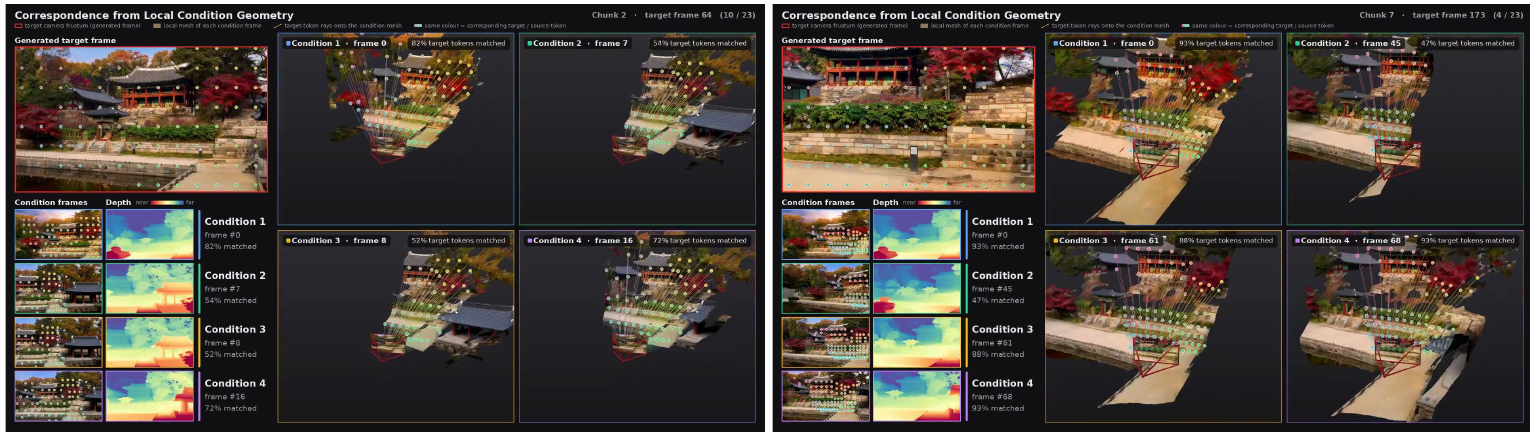}
    \caption{
        \textbf{Visualization of Geometry-Guided Patch Correspondence Construction.} Local geometry from each historical observation is used to build a history-to-target patch correspondence cache at latent resolution. Guided by this cache, GCA allows each noisy target token to attend only to its geometrically matched historical memory tokens as keys and values.
    }
    \label{fig:appendix:patch_memory}
\end{figure*}

\vspace{-1.25 em}
\paragraph{Dual-view face-ID rasterization.}
As shown in Fig.~\ref{fig:appendix:patch_memory}, for each selected source frame $s$ and target frame $t$, we transform the vertices of the \emph{same} local mesh $G_s$ into the respective
camera clip spaces and rasterize them with \texttt{nvdiffrast}~\citep{Laine2020diffrast} at the
latent resolution $H_\ell\times W_\ell$:
\begin{equation}
    F_s=\mathcal{R}(G_s;c_s),\qquad
    F_{s\rightarrow t}=\mathcal{R}(G_s;c_t).
    \label{eq:supp_dual_rasterization}
\end{equation}
The rasterizer's depth test assigns each covered latent-grid location
the ID of its nearest visible triangle; a zero ID denotes background.
We use these discrete IDs without interpolating them and express both
maps in a common image-coordinate convention before matching.

Because the two maps refer to the same source mesh, a valid shared
face ID defines a source--target patch correspondence:
\begin{equation}
    \mathcal{C}_{s\rightarrow t}
    =\bigl\{(i,j)\ \big|\ F_{s\rightarrow t}(i)
       =F_s(j)>0\bigr\},
    \label{eq:supp_face_correspondence}
\end{equation}
where $i$ and $j$ index target and source latent-grid locations,
respectively. Face IDs are matched only \emph{within} each source
mesh; the source-frame index is retained when combining matches from
multiple historical views. The resulting cache stores the matched
source-patch coordinates for each target patch and source frame, with
unmatched entries marked invalid. These per-source candidates are
subsequently filtered using the target-view Invisible Octree mask
before they are accessed by GCA.
\begin{figure*}[t]
    \centering
    \scriptsize
    \includegraphics[width=\textwidth]{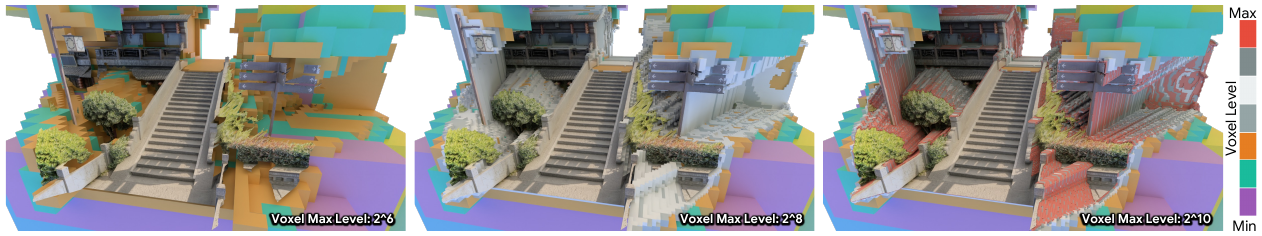}
    \caption{
        \textbf{Visualization of the Invisible Octree update process.} Partially visible voxels are recursively subdivided until each node becomes fully visible or fully invisible, or the maximum voxel resolution is reached. With increasing resolution, the Invisible Octree progressively approximates the true invisible regions in the target view.
    }
    \label{fig:appendix:vis_octree_level}
\end{figure*}

\subsection{Invisible Octree Construction and Streaming Update}
\label{sec:appendix:additional_method_details:invisible_octree}

\paragraph{Octree states.}
We maintain an Invisible Octree $\mathcal{O}$ as a sparse proxy for
accumulated visibility evidence along the camera trajectory.
Each allocated node $v$ is assigned one of three
states: \emph{free}, \emph{invisible}, or \emph{partially visible}.
A free node lies entirely in observed free space along the relevant
camera rays, whereas an invisible node lies entirely behind the
observed depth surface and has not yet been resolved by an observation.
A partially visible node intersects the boundary between these
regions, or contains a mixture of visible and invisible space.

Given a camera with depth map $D$, let $[z_v^{-}, z_v^{+}]$ be the depth range
of $v$ in camera space, and let $d_v^{\min}$ and $d_v^{\max}$ be the
minimum and maximum valid depths over its projected footprint. A node containing no observed surface points is classified as
\begin{equation}
    s(v) =
    \begin{cases}
        \text{free}, & \begin{array}[t]{@{}l@{}}
                           z_v^{+} < d_v^{\min}\ \text{or no valid} \\
                           \text{depth in the footprint},
                       \end{array} \\
        \text{invisible}, & z_v^{-} > d_v^{\max}, \\
        \text{partially visible}, & \text{otherwise}.
    \end{cases}
    \label{eq:supp_octree_classify}
\end{equation}
Free and invisible nodes are terminal for the current visibility update.
Partially visible nodes are recursively subdivided until their
children can be classified or the maximum resolution is reached.
Only invisible leaf nodes can be revisited using subsequent camera observations.

\vspace{-1.25 em}
\paragraph{Initialization from the first view.}
Given the first camera $c_1$ and depth map $D_1$, we set the octree's
world-space bounds according to the scene scale and initialize a
coarse grid with $N_{\min}=2^4$ cells per axis. A one-cell-thick \emph{background shell} of invisible cells encloses the full scene to account for regions with invalid depth estimates.

Both the global invisible octree $\mathcal{O}$ and a global visible mesh
$\mathcal{G}^{\mathrm{vis}}$, triangulated from back-projected depth,
are initialized from the first frame.
Octree nodes are classified against depths rendered from $\mathcal{G}^{\mathrm{vis}}$, ensuring consistency between the two global proxies.
For the first chunk, classification starts from the coarsest grid and recursively refines the octree up to a resolution equivalent to $2^{10}$ cells per axis.
The two structures serve complementary purposes:
$\mathcal{O}$ represents currently unresolved invisible space, whereas $\mathcal{G}^{\mathrm{vis}}$ represents surfaces that have already been observed.
\textbf{The global visible mesh is used only for visibility comparison;} the source-specific local meshes used to construct GCA correspondences remain independent.
We visualize the octree subdivision and update process in Fig.~\ref{fig:appendix:vis_octree_level}.

\vspace{-1.25 em}
\paragraph{Identifying regions to update.}
For a newly generated frame with camera $c_t$ and estimated depth
$D_t$, we first render both global proxies into its view.
We build a BVH over the invisible octree leaves and ray-cast it to obtain the
depth $d_t^{\mathrm{inv}}(p)$ of the first invisible voxel along each
camera ray. Separately, we rasterize $\mathcal{G}^{\mathrm{vis}}$ to
obtain its visible-surface depth $d_t^{\mathrm{vis}}(p)$.
Their depth ordering identifies image regions where previously invisible space appears in front of the surface already represented by the global visible mesh.
With a small comparison tolerance $\epsilon$, the corresponding update mask
can be written as
\begin{equation}
    M_t^{\mathrm{upd}}(p)
    =
    \mathbb{1}\!\left[
        d_t^{\mathrm{inv}}(p) + \epsilon
        < d_t^{\mathrm{vis}}(p)
    \right].
    \label{eq:supp_octree_update_mask}
\end{equation}
Only pixels with a valid current depth and a valid invisible-voxel
intersection are considered for this comparison.
We use $D_t$ within $M_t^{\mathrm{upd}}$ to reclassify the
corresponding octree region by the same depth-based procedure used during initialization.
Previously invisible nodes can therefore be refined as new observations reveal their contents.
We further back-project and triangulate the masked depth observations and incorporate the resulting surfaces into $\mathcal{G}^{\mathrm{vis}}$.
Restricting both updates to newly exposed regions prevents established visibility evidence from being repeatedly overwritten by depth estimates from later generated frames.

\vspace{-1.25 em}
\paragraph{Visibility-aware correspondence filtering.}
The global Invisible Octree and visible mesh are updated after each generated chunk as the camera progresses along its trajectory.
For each target camera, we project the accumulated global proxies into the target view and compare their depths to determine the target-view invisible mask.
This mask is then applied directly to the source--target correspondence cache, removing candidates that fall within regions determined to be occluded before GCA accesses the corresponding historical tokens.
Importantly, neither global proxy provides appearance features to the generator.
Appearance information remains entirely within the historical frame latents, while $\mathcal{O}$ and $\mathcal{G}^{\mathrm{vis}}$ provide only visibility information for filtering their geometrically addressed correspondences.
\subsection{Keyframe History Retrieval}
\label{sec:appendix:additional_method_details:keyframe_retrieval}

For each target chunk, we retrieve historical frames based on their
coverage of the views to be generated. Let $\mathcal{T}$ denote the
target frames in the chunk, and let
$\mathcal{Q}=\{(u,i)\mid u\in\mathcal{T},\,i\in\Omega_u\}$ be the set
of their spatial locations. For each historical frame $s$, we project
its depth-derived local geometry into every target view. After
filtering occluded projections with the target-view Invisible Octree
mask, we obtain a coverage set $\mathcal{V}_s\subseteq\mathcal{Q}$.

We retain the first and latest historical frames to provide scene
identity and continuity across chunks. Starting from these frames,
we greedily add the candidate with the largest marginal contribution
to target-view coverage. To limit redundant selection from densely
observed regions, we track how many selected frames cover each target
location $q$:
\begin{equation}
\begin{aligned}
    n_q(\mathcal{S})
    &= \sum_{s\in\mathcal{S}}
      \mathbb{1}[q\in\mathcal{V}_s],
    \\
    \Delta(s\mid\mathcal{S})
    &= \sum_{q\in\mathcal{V}_s}
      \mathbb{1}[n_q(\mathcal{S})<N_{\mathrm{covered}}],
\end{aligned}
    \label{eq:supp_history_coverage}
\end{equation}
where $\mathcal{S}$ is the current set of selected frames and
$N_{\mathrm{covered}}=3$. At each iteration, we select the remaining
frame with the largest $\Delta(s\mid\mathcal{S})$ until the history
budget is reached. Once a target location has been covered three
times, further coverage of that location contributes no additional
score. This encourages the selected conditions to span the upcoming
target views while retaining multiple observations where available.
\subsection{Depth Update for Streaming Generation}
\label{sec:appendix:additional_method_details:depth_update}

After generating each video chunk, we estimate the depths of its
decoded frames using Depth Anything 3~\citep{depthanything3}. Estimating the new frames
in isolation could introduce inconsistencies with the geometry stored
in the history bank. We therefore include uniformly sampled
historical frames as anchors. For a history bank containing $N_{\mathrm{hist}}$ frames, we use the sampling
interval $N_{\mathrm{interval}}
    = \max\!\left(1,
    \left\lfloor\frac{N_{\mathrm{hist}}}{25}\right\rfloor\right)$
and select historical frames at this interval in temporal order.
We jointly feed the anchor frames and newly generated frames to DA3,
together with their corresponding camera intrinsics and extrinsics.
We use its pose-conditioned mode so that depth estimation is informed
by the prescribed camera trajectory. We retain only
the depths of the newly generated frames; the depths already stored
in the history bank are left unchanged. These new depth maps are then
appended to the bank and used to update the visibility proxies for
subsequent chunks.

\begin{table}[t]
    \centering
    \vspace{1.0 em}
    \caption{\textbf{Training configuration.}
        Rank-32 LoRA adapters and GCA modules are trained jointly on the frozen Wan2.1-I2V-14B backbone.}
    \label{tab:supp_training_config}
    \small
    \renewcommand{\arraystretch}{1.18}
    \setlength{\tabcolsep}{12pt}
    \begin{adjustbox}{max width=\linewidth}
    \begin{tabular}{@{}l@{\hspace{2.6em}}l@{}}
        \specialrule{.15em}{.1em}{.1em}
        \textbf{Setting} & \textbf{Value} \\
        \midrule
        Backbone & Wan2.1-I2V-14B~\citep{wan2025wan} \\
        Training resolution & $480\times832$ \\
        LoRA rank & 32 \\
        GCA hidden dimension & 640 \\
        GCA placement & Even-indexed DiT blocks \\
        Numerical precision & BF16 mixed precision \\
        I2V / H2V sampling ratio & 30\% / 70\% \\
        Retrieved history keyframes & 9 \\
        Optimizer & AdamW \\
        Peak learning rate & $10^{-4}$ \\
        Learning-rate warm-up & 1K iterations \\
        Training iterations & 10K \\
        Global batch size & 32 \\
        Hardware & 32 GPUs \\
        \specialrule{.15em}{.1em}{.1em}
    \end{tabular}
    \end{adjustbox}
    \vspace{1.0 em}
\end{table}

\begin{table}[t]
    \centering
    \caption{\textbf{Inference cost.}
        Denoising time is measured per step under matched inputs.}
    \label{tab:supp_runtime}
    \small
    \renewcommand{\arraystretch}{1.18}
    \setlength{\tabcolsep}{10pt}
    \begin{adjustbox}{max width=\linewidth}
    \begin{tabular}{@{}l@{\hspace{1.8em}}l@{\hspace{1.8em}}c@{}}
        \specialrule{.15em}{.1em}{.1em}
        \textbf{Operation} & \textbf{Scope} & \textbf{Cost} \\
        \midrule
        Wan2.1-14B & Per step, One GPU & 33.8\,s / 39.9\,GB \\
        Wan2.1-14B with GCA & Per step, One GPU & 34.5\,s \, (+2.1\%) / 40.9\,GB \\
        Depth Anything 3 & 48-frame call & 14\,s \\
        Invisible Octree update & 23-frame chunk & 3.4\,s \\
        \specialrule{.15em}{.1em}{.1em}
    \end{tabular}
    \end{adjustbox}
\end{table}

\section{Dataset and Implementation Details}
\label{sec:appendix:dataset_details}

\paragraph{Dataset preprocessing.}
We train on DL3DV-10K~\citep{ling2024dl3dv} after filtering scenes with pronounced motion
blur or insufficient illumination, retaining approximately 6.5K
scenes. Each retained sequence is divided into consecutive,
non-overlapping 55-frame clips at a resolution of $480\times832$.
We use Depth Anything 3~\citep{depthanything3} to obtain per-frame camera poses and depths,
and Qwen3-VL-8B-Instruct~\citep{bai2025qwen3} to generate video captions. For
history-to-video (H2V) training, the first 32 frames form the history
bank, from which nine keyframes are retrieved; the remaining 23
frames serve as generation targets. After filtering and processing, the resulting dataset comprises approximately 30K high-quality video clips.

\vspace{-1.25 em}
\paragraph{Training configuration.}
We use Wan2.1-I2V-14B~\citep{wan2025wan} as the pretrained backbone and freeze its
original parameters. Rank-32 LoRA adapters and GCA modules are
jointly trained, with GCA inserted after self-attention in every
even-indexed DiT block. Each GCA module has a hidden dimension of
640. We zero-initialize its output projection $W_O$, so that the GCA
residual is initially zero and does not perturb the pretrained
backbone features at the start of training. The complete training settings
are summarized in Table~\ref{tab:supp_training_config}.

\vspace{-1.25 em}
\paragraph{Inference configuration and computation cost.}
We use 25 denoising steps with a classifier-free guidance scale of 5. Table~\ref{tab:supp_runtime} reports the denoising time per step, depth-estimation latency, and Invisible Octree update time. We measure backbone-only and GCA-enhanced denoising on a single GPU using identical input dimensions and history configurations. Depth-estimation latency is measured for a 48-frame inference call, while Invisible Octree update time is reported per 23-frame generated chunk. GCA adds only 1.9\% to the backbone parameter count with minimal denoising overhead. We implement Invisible Octree management and updates in Warp~\citep{macklin2022warp} to enable parallel execution on the GPU.
Both coverage-based history retrieval and rasterization-based correspondence construction can be efficiently parallelized on the GPU, introducing negligible computational overhead during inference.

\section{Evaluation Protocols}
\label{sec:appendix:evaluation_protocols}

\subsection{Baseline Configuration and Method Comparison}
\label{sec:appendix:evaluation_protocols:baseline_configuration_and_method_comparison}

\paragraph{Baseline availability.}
We discuss AnchorWeave~\citep{wang2026anchorweave} as a related method, but exclude it from quantitative comparisons since its inference weights were not publicly available at the time of evaluation.

\vspace{-1.25 em}
\paragraph{Depth and camera-scale alignment.}
DL3DV-Evaluation~\citep{ling2024dl3dv} provides camera trajectories whose scale may differ from that of the depth predicted by baselines. Such a mismatch changes the effective magnitude of the prescribed camera motion and can confound camera-control comparisons. To establish a common geometric scale, we estimate multi-view depths for each evaluation scene using Depth Anything 3~\citep{depthanything3}. For the explicit memory methods requiring an initial-frame depth map, we align their predicted depth to the DA3 estimate of the first frame by least-squares scale fitting over valid pixels.
We also provide the same aligned initial-frame depth to the geometry-based baselines. For UCM~\citep{xu2026ucm}, we replace the depth estimator used in its original implementation with pose-conditioned DA3, as used by Lyra 2.0~\citep{shen2026lyra} and \method in our evaluation, because depth estimated without conditioning on the supplied camera poses may be inconsistent with the evaluation trajectory.

\vspace{-1.25 em}
\paragraph{Implicit-memory baselines.}
For HY-WorldPlay~\citep{sun2025worldplay}, Lingbot-World~\citep{team2026advancing}, and Infinite-World~\citep{wu2026infiniteworld}, we normalize camera translations according to scene scale before inference while preserving the prescribed camera rotations and relative motion.
Infinite-World~\citep{wu2026infiniteworld} accepts discrete action inputs rather than continuous camera poses; following its processing protocol, we convert consecutive relative camera transformations into the corresponding action sequence. Its trajectory-adherence results should therefore be interpreted in light of this action discretization.

\begin{figure*}[t]
    \centering
    \scriptsize
    \vspace{-1.5 em}
    \includegraphics[width=\textwidth]{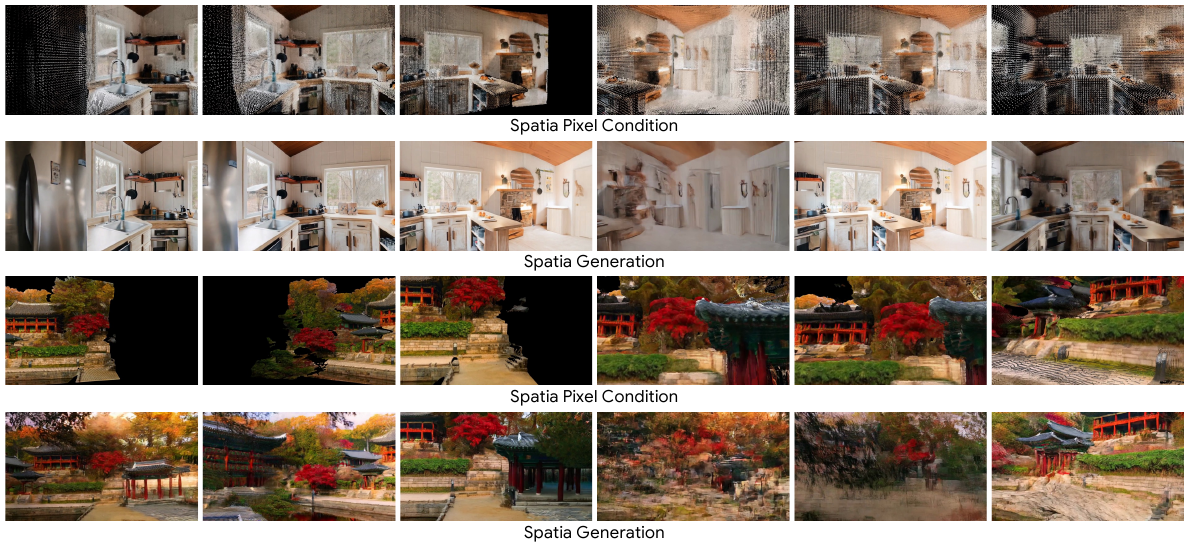}
    \caption{
        \textbf{Visualization of the 3D pixel condition and generated frames of Spatia.}
        Even with relatively clean 3D pixel-aligned conditioning, Spatia exhibits visible temporal instability and image degradation. Under more complex camera trajectories and noisier 3D pixel-aligned conditioning, frame jitter emerges in the first generated chunk, followed by a complete breakdown of visual content in the second.
    }
    \label{fig:appendix:spatial_problem}
\end{figure*}

\begin{figure*}[t]
    \centering
    \scriptsize
    \includegraphics[width=\textwidth]{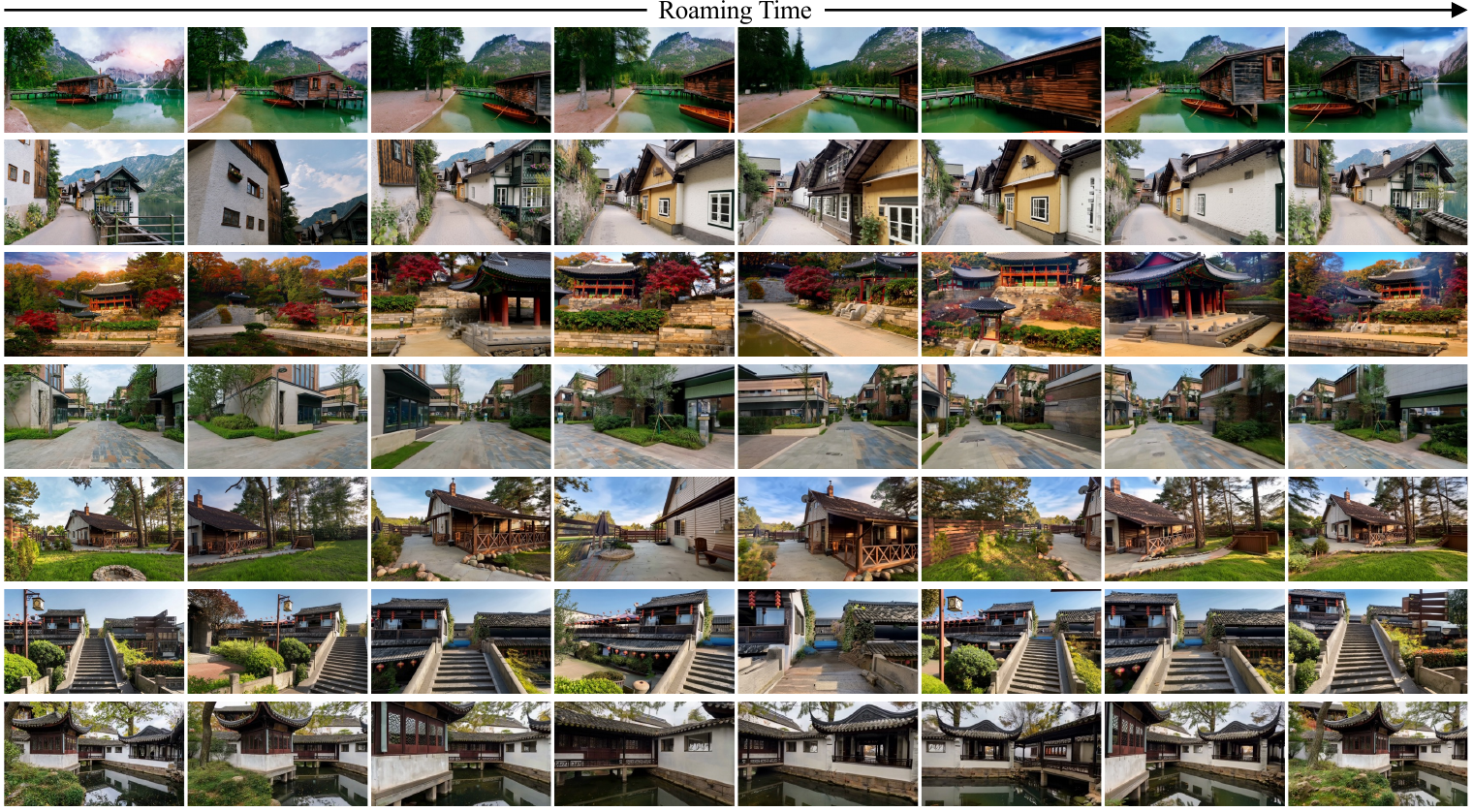}
    \caption{
        \textbf{More results across a broader range of data.}
    }
    \label{fig:appendix:more_case}
\end{figure*}

\vspace{-1.25 em}
\paragraph{Geometry-based correspondence methods.}
Similar to our method, UCM~\citep{xu2026ucm} and Lyra 2.0~\citep{shen2026lyra} leverage estimated geometry to establish correspondences between historical observations and target viewpoints, without fusing these observations into a persistent global 3D representation. The key distinction lies in how the resulting correspondences are incorporated into the generation process.
Following PE-Field~\citep{bai2025positional}, UCM projects historical observations into relevant target views and warps their positional encodings, allowing historical and target tokens to interact through geometry-aware attention. To limit computation, each historical frame is assigned a single relevant target viewpoint for this warping. Lyra 2.0 instead forward-warps canonical source coordinates and depth from multiple retrieved frames, encodes the resulting correspondence maps, and adds their embeddings to DiT tokens. These designs provide geometric cues for memory access, but do not explicitly restrict each target patch to attend to its set of matched historical patches.

\method stores the source coordinates of valid history-to-target patch correspondences and uses them to gather historical features as keys and values for Geometric Correspondence Attention. A target patch can draw on multiple historical observations, while the Invisible Octree filters candidates that are projectable but occluded in the target view. The resulting attention output is injected through a residual branch during denoising. This provides direct access to the visual content of geometrically matched memory patches while keeping depth errors local to individual source views.

In the comparisons in Figs.~\ref{fig:exp:dl3dv_comparison} and~\ref{fig:exp:long_horizon_comparison}, UCM preserves coherent views early in the rollout but exhibits increasing camera drift and visual degradation under longer or faster camera motion. Lyra 2.0 is the strongest geometry-correspondence baseline in these examples, yet also degrades under substantial viewpoint changes. These observations are consistent with the quantitative results in Table~\ref{tab:quantitative_comparison}.

\vspace{-1.25 em}
\paragraph{Globally fused 3D memory methods.}
Spatia~\citep{zhao2026spatia} reconstructs historical observations with MapAnything~\citep{keetha2026mapanything}, updates a persistent scene point cloud, and renders it from target viewpoints to produce spatial guidance for subsequent video generation. As shown in Fig.~\ref{fig:appendix:spatial_problem}, this can recover convincing observations on some relatively simple rotational trajectories in WorldScore-Static. In the more challenging DL3DV-Evaluation examples and long-horizon trajectories, however, we observe scene distortion and loss of previously visible content, in some cases beginning within the first generation window and becoming more severe in subsequent rollouts. These failures are consistent with errors in the accumulated point cloud being repeatedly rendered into the conditioning signal. \method avoids this source of persistent geometric error by retaining visual observations as frame latents and using their independently estimated geometry only to address memory.

\section{Additional Qualitative and Video Results}
\label{sec:appendix:more_results}

\begin{figure*}[t]
    \centering
    \scriptsize
    \vspace{-6 em}
    \includegraphics[width=0.9\textwidth]{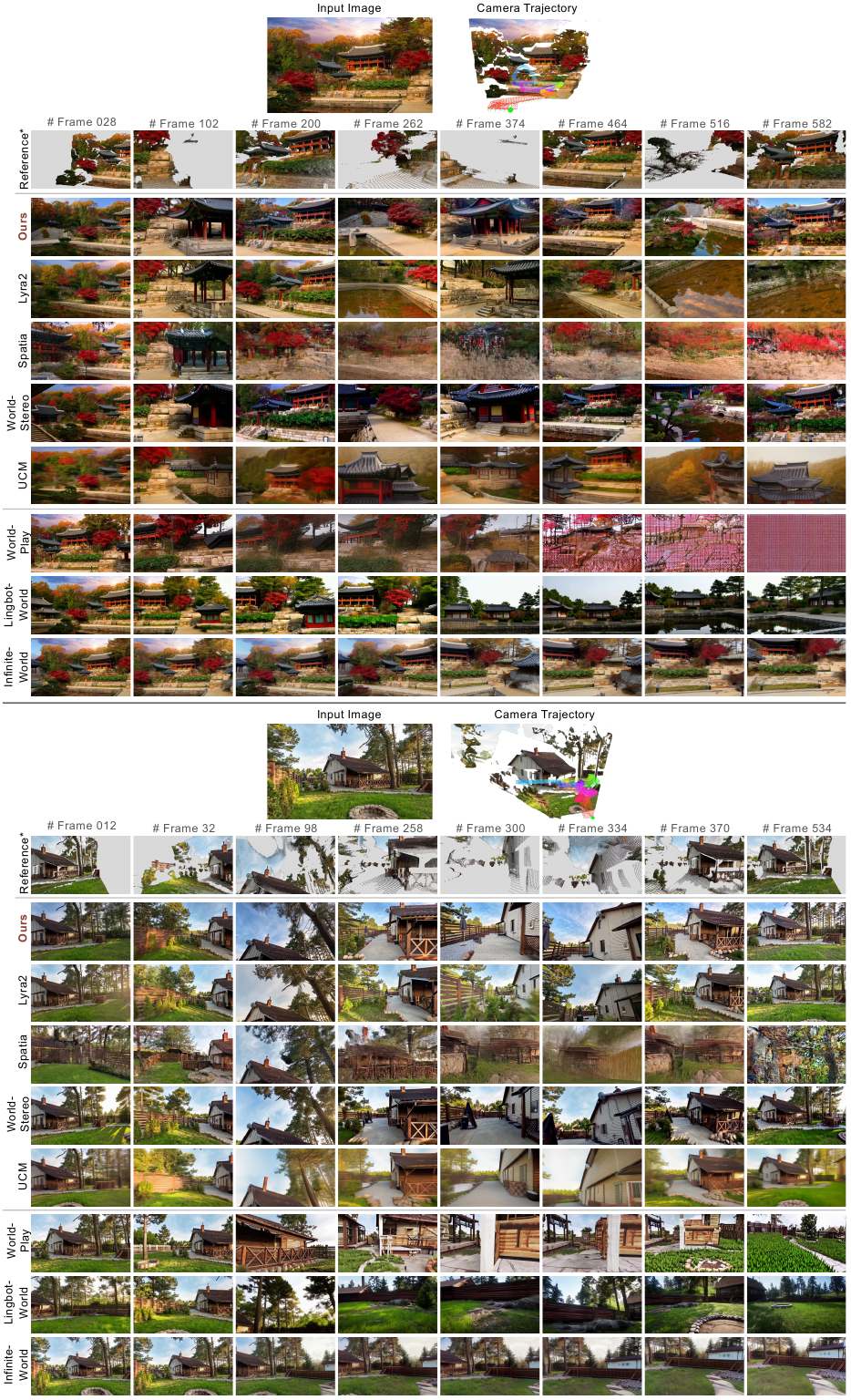}
    \caption{
        \textbf{Qualitative comparison of minute-long challenging camera trajectories results}
    }
    \label{fig:appendix:long_horizon_comparison}
\end{figure*}

\begin{figure*}[t]
    \centering
    \scriptsize
    \vspace{-7 em}
    \includegraphics[width=0.8\textwidth]{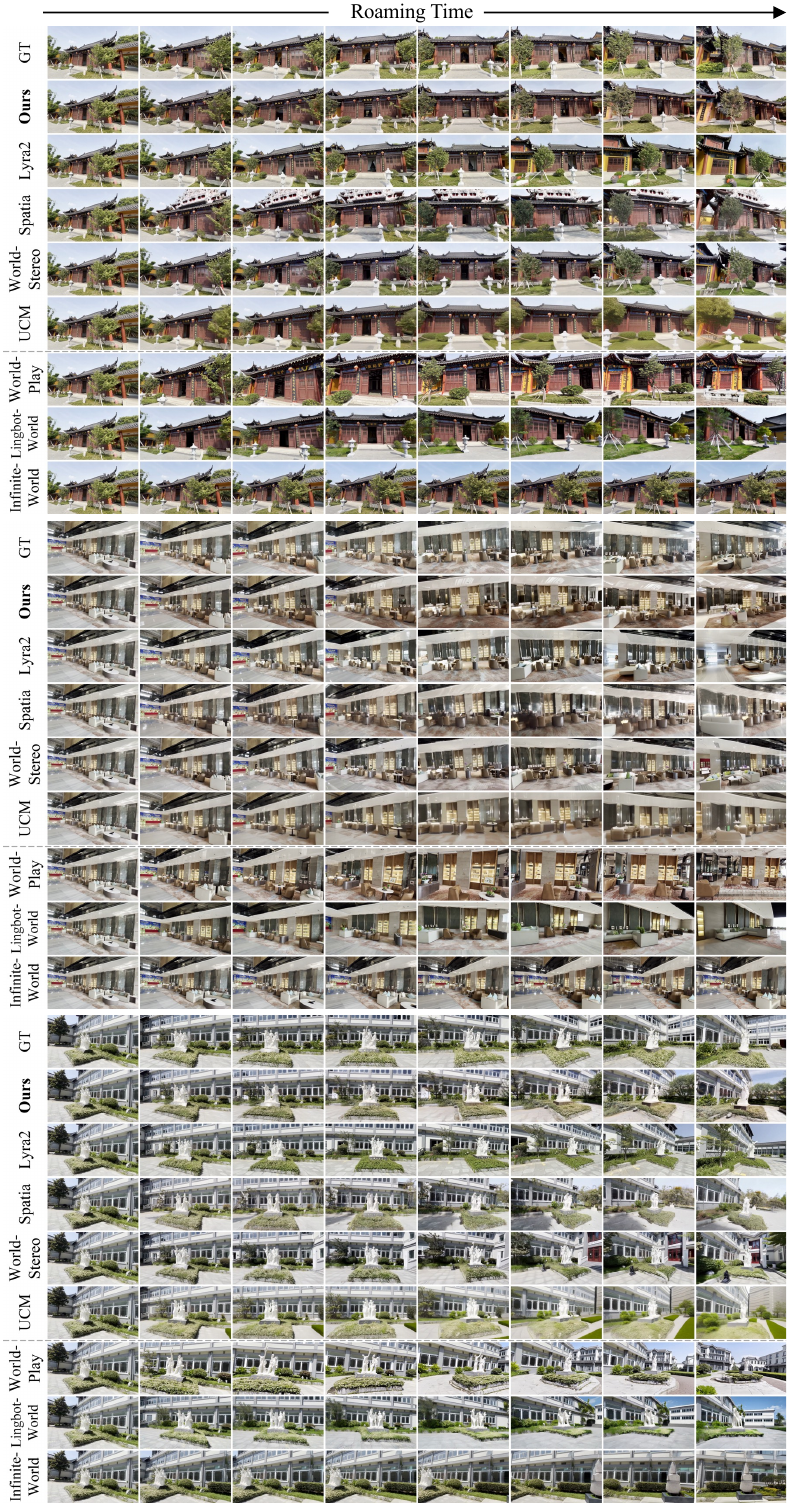}
    \caption{
        \textbf{Qualitative comparison of DL3DV-Evaluation results.}
    }
    \label{fig:appendix:dl3dv_comparison_0}
\end{figure*}

\begin{figure*}[t]
    \centering
    \scriptsize
    \vspace{-7 em}
    \includegraphics[width=0.8\textwidth]{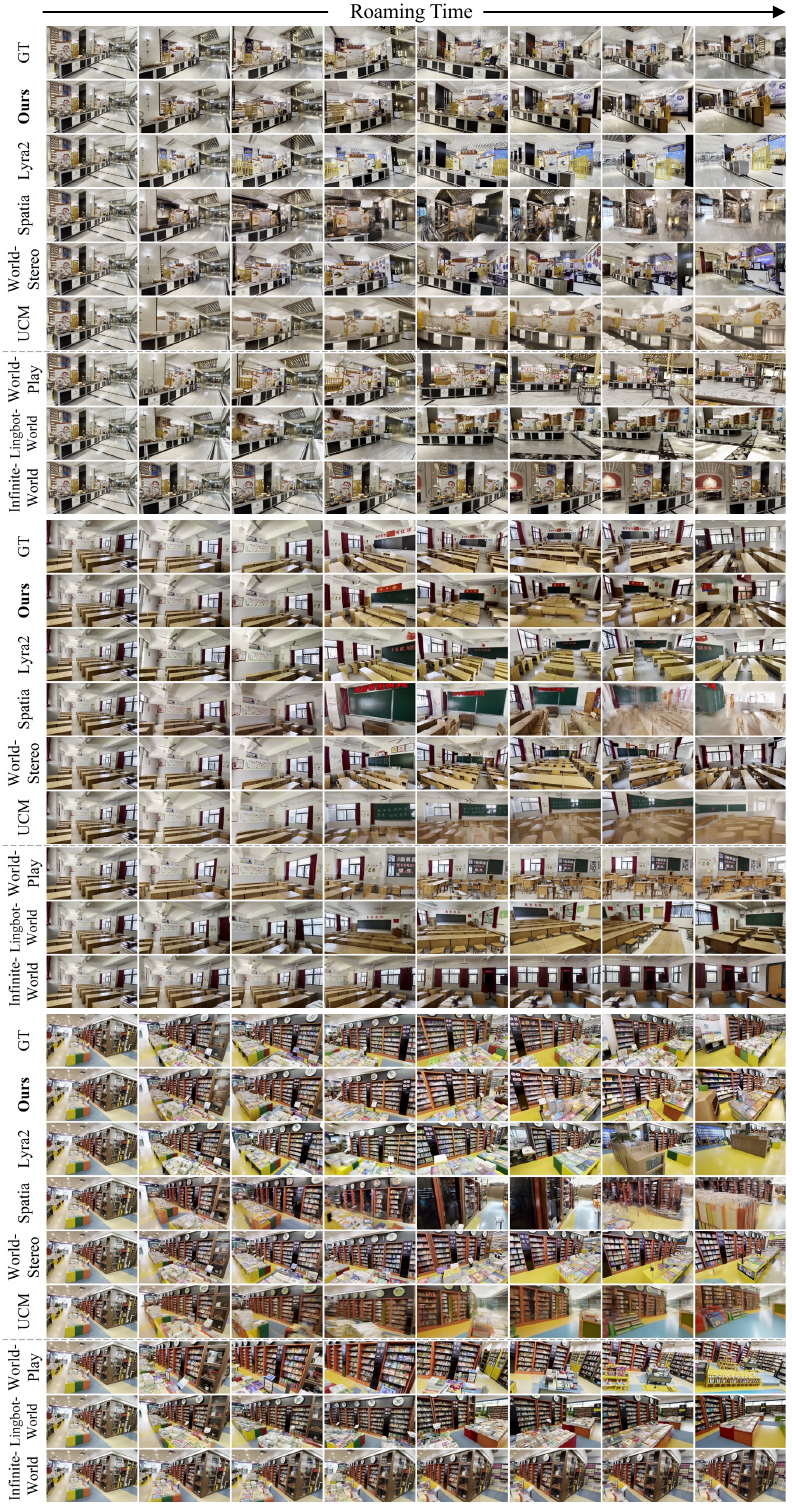}
    \caption{
        \textbf{Qualitative comparison of DL3DV-Evaluation results.}
    }
    \label{fig:appendix:dl3dv_comparison_1}
\end{figure*}

\begin{figure*}[t]
    \centering
    \scriptsize
    \vspace{-7 em}
    \includegraphics[width=0.8\textwidth]{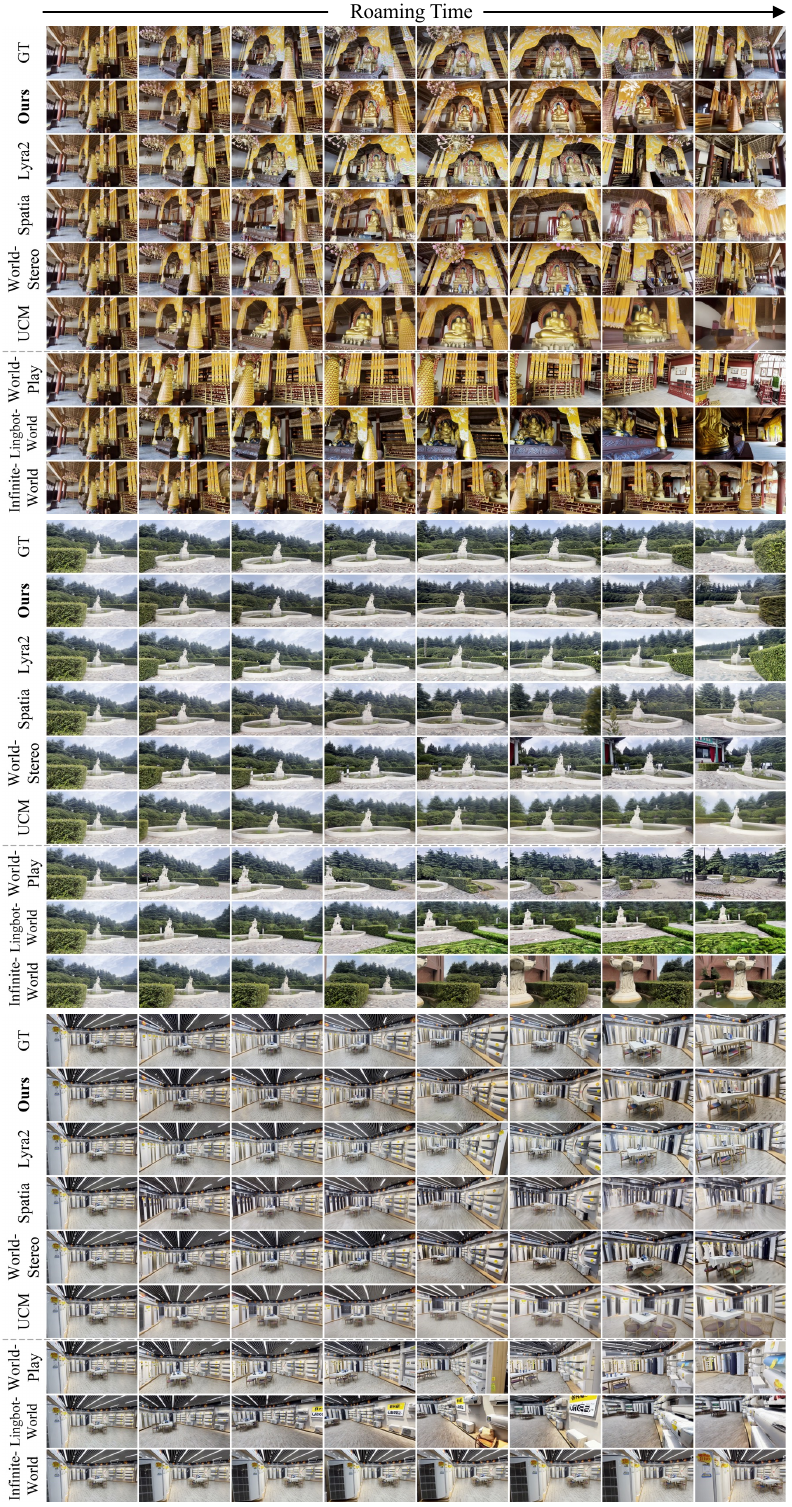}
    \caption{
        \textbf{Qualitative comparison of DL3DV-Evaluation results.}
    }
    \label{fig:appendix:dl3dv_comparison_2}
\end{figure*}

\begin{figure*}[t]
    \centering
    \scriptsize
    \vspace{-7 em}
    \includegraphics[width=0.775\textwidth]{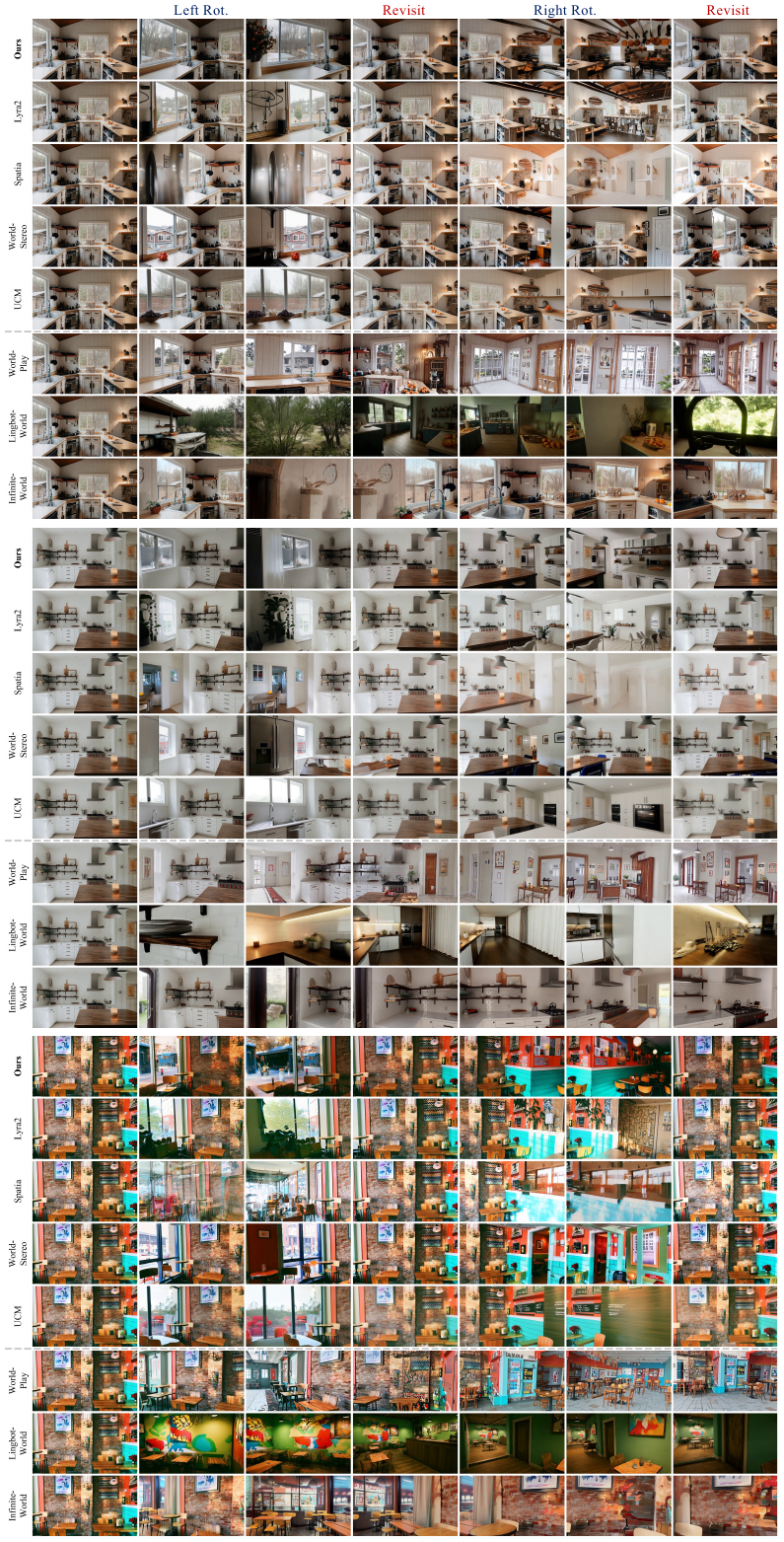}
    \caption{
        \textbf{Qualitative comparison of WorldScore-Static results.}
    }
    \label{fig:appendix:worldscore_comparison_0}
\end{figure*}

\begin{figure*}[t]
    \centering
    \scriptsize
    \vspace{-7 em}
    \includegraphics[width=0.775\textwidth]{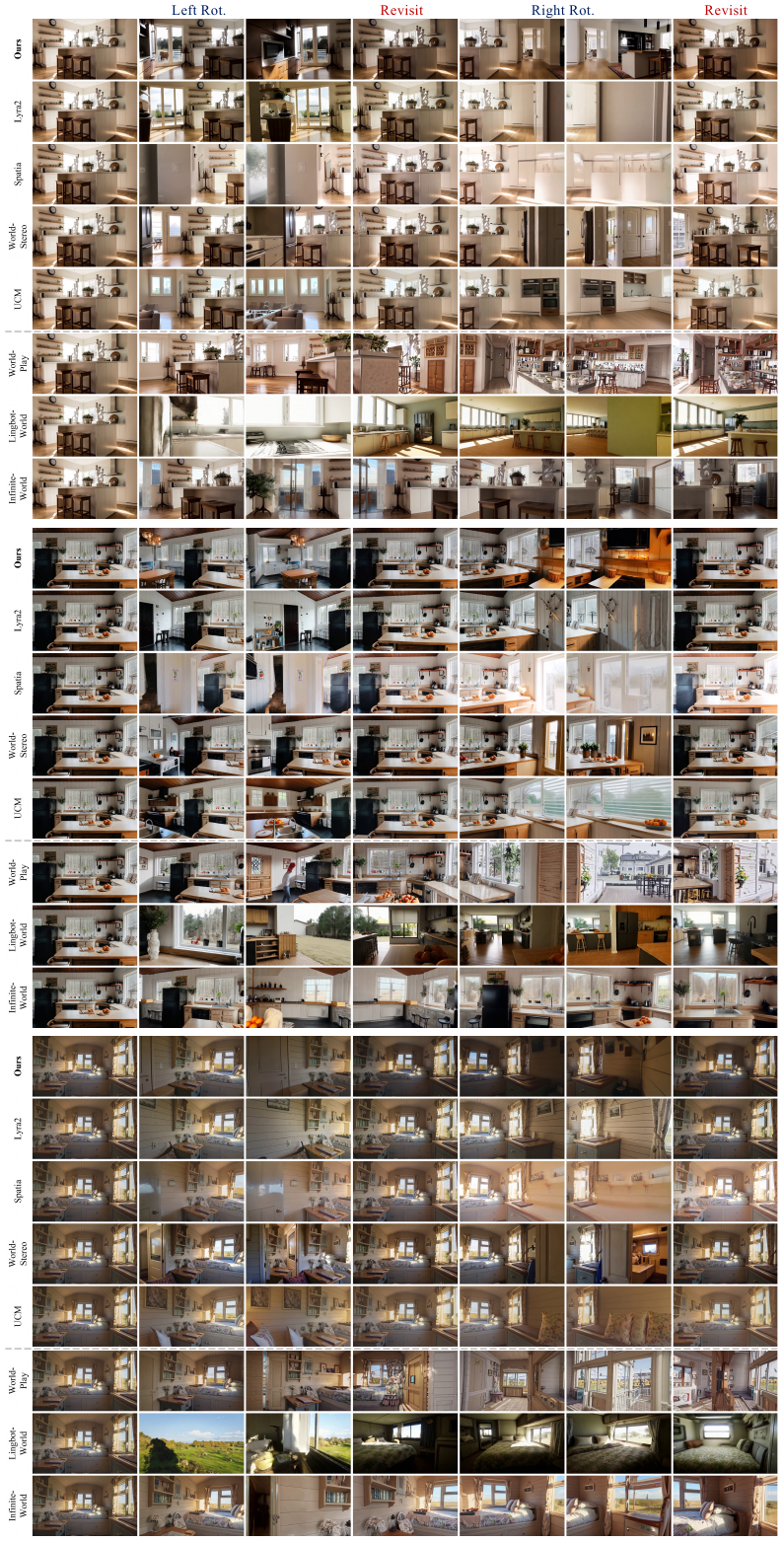}
    \caption{
        \textbf{Qualitative comparison of WorldScore-Static results.}
    }
    \label{fig:appendix:worldscore_comparison_1}
\end{figure*}

\begin{figure*}[t]
    \centering
    \scriptsize
    \vspace{-7 em}
    \includegraphics[width=0.775\textwidth]{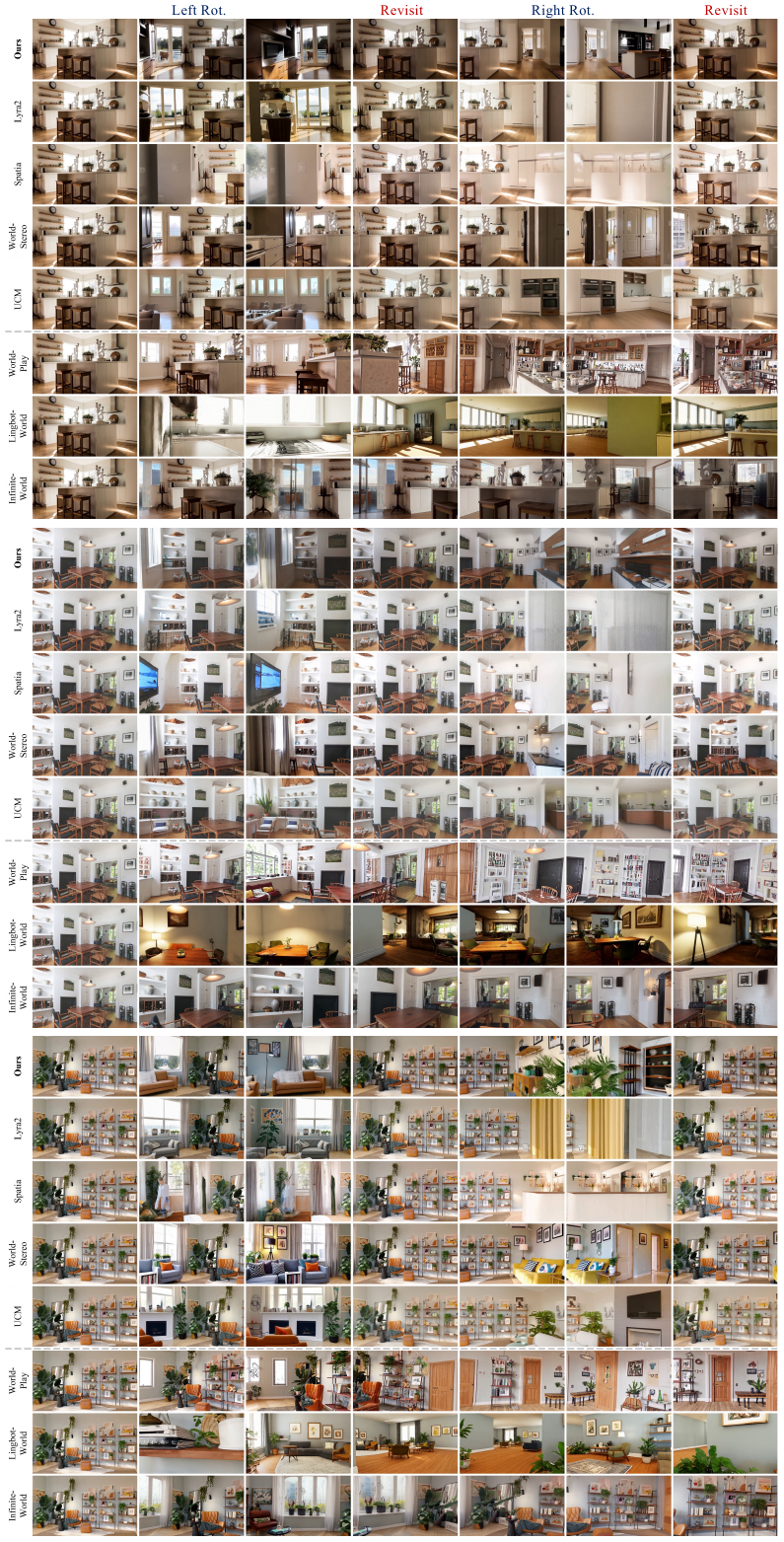}
    \caption{
        \textbf{Qualitative comparison of WorldScore-Static results.}
    }
    \label{fig:appendix:worldscore_comparison_2}
\end{figure*}

We present additional qualitative results on DL3DV-Evaluation (Fig.~\ref{fig:appendix:dl3dv_comparison_0},~\ref{fig:appendix:dl3dv_comparison_1},~\ref{fig:appendix:dl3dv_comparison_2}), WorldScore-Static (Fig.~\ref{fig:appendix:worldscore_comparison_0},~\ref{fig:appendix:worldscore_comparison_1},~\ref{fig:appendix:worldscore_comparison_2}), and the Long-Horizon dataset (Fig.~\ref{fig:appendix:long_horizon_comparison}), together with further results demonstrating the performance of our method across a broader range of data (Fig.~\ref{fig:appendix:more_case}).
\textbf{Please refer to our project page: \projectpage{} for richer and more dynamic visualizations.}

\clearpage

\end{document}